%% file: main.tex
\documentclass[10pt, a4paper]{article}
\usepackage{hybridreport}

\usepackage{array}
\usepackage{multirow}
\usepackage{wrapfig}
\usepackage{adjustbox}
\usepackage[round]{natbib}

\graphicspath{{./}}

\newcommand{\specifymorename}{\textbf{Specify More}}
\newcommand{\seelessname}{\textbf{See Less}}
\newcommand{\stwoname}{S2}
\newcommand{\stdcell}[2]{\mbox{#1 {\color[gray]{0.45}\fontsize{5}{5}\selectfont$\pm$#2}}}

\newcommand\blfootnote[1]{%
  \begingroup\renewcommand\thefootnote{}\footnote{#1}\addtocounter{footnote}{-1}\endgroup}

\defcitealias{physicalintelligence2025pi05}{Black et al., 2025}

\reporttype{Technical Report}
\logo{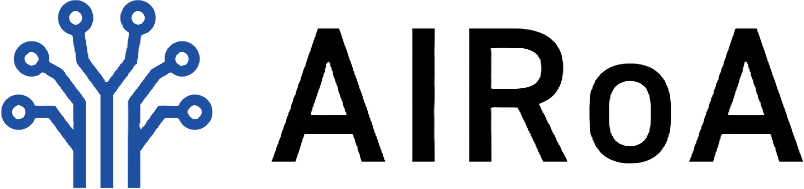}
\title{See Less, Specify More: Visual Evidence Budgets for Generalizable VLAs}
\shorttitle{See Less, Specify More: Visual Evidence Budgets for Generalizable VLAs}
\author{%
  Yueh-Hua (Kris) Wu\textsuperscript{*}\quad
  Tatsuya Matsushima\quad
  Kei Ota%
}
\institution{AIRoA}
\date{May 2026}
\correspondence{}

\abstracttext{%
  Generalization remains a central bottleneck for vision-language-action (VLA)
  models: under distractors, appearance shifts, and semantically similar tasks,
  the policy must often infer local execution details from coarse instructions
  while also deciding which parts of the image matter for control. We present
  \stwoname{} (\seelessname{}, \specifymorename{}), a framework for improving VLA
  generalization by training the executor under a cleaner interface.
  \specifymorename{} preserves the original instruction as a stable high-level
  goal while relabeling each trajectory into refined trajectory- and
  subtask-level language that disambiguates the current execution mode. Unlike
  native attention, \seelessname{} imposes an explicit visual evidence budget,
  training the executor to act from task-sufficient evidence rather than
  unconstrained visual context, without any region or mask annotation. This
  interface lets the executor follow detailed guidance without relying on
  distracting visual patches or resolving avoidable ambiguity on its own, and it
  remains compatible with off-the-shelf VLM planners through in-context learning.
  Across our main evaluation settings, S2 improves overall generalization
  metrics by changing the executor's learning problem: coarse instructions
  induce avoidable supervision aliasing, goal-preserving local guidance
  outperforms instruction replacement in our main ablations, and explicit
  evidence budgeting reduces dependence on broad visual context beyond
  efficiency considerations. Across eight real-robot tasks on
  TX-G2 (an AgiBot G2-compatible variant) and HSR, S2 raises mean subtask
  success from 54.2\% to 79.0\% over $\pi_{0.5}$. Together, these results suggest
  that VLA generalization improves when the executor is trained to act from
  informative local guidance and task-sufficient visual evidence, rather than
  recovering both from weak supervision.
}

\begin{document}
\maketitle
\blfootnote{\textsuperscript{*}Corresponding author: \href{mailto:kris.wu@airoa.org}{\texttt{kris.wu@airoa.org}}.}

\begin{center}
  \includegraphics[width=0.9\textwidth]{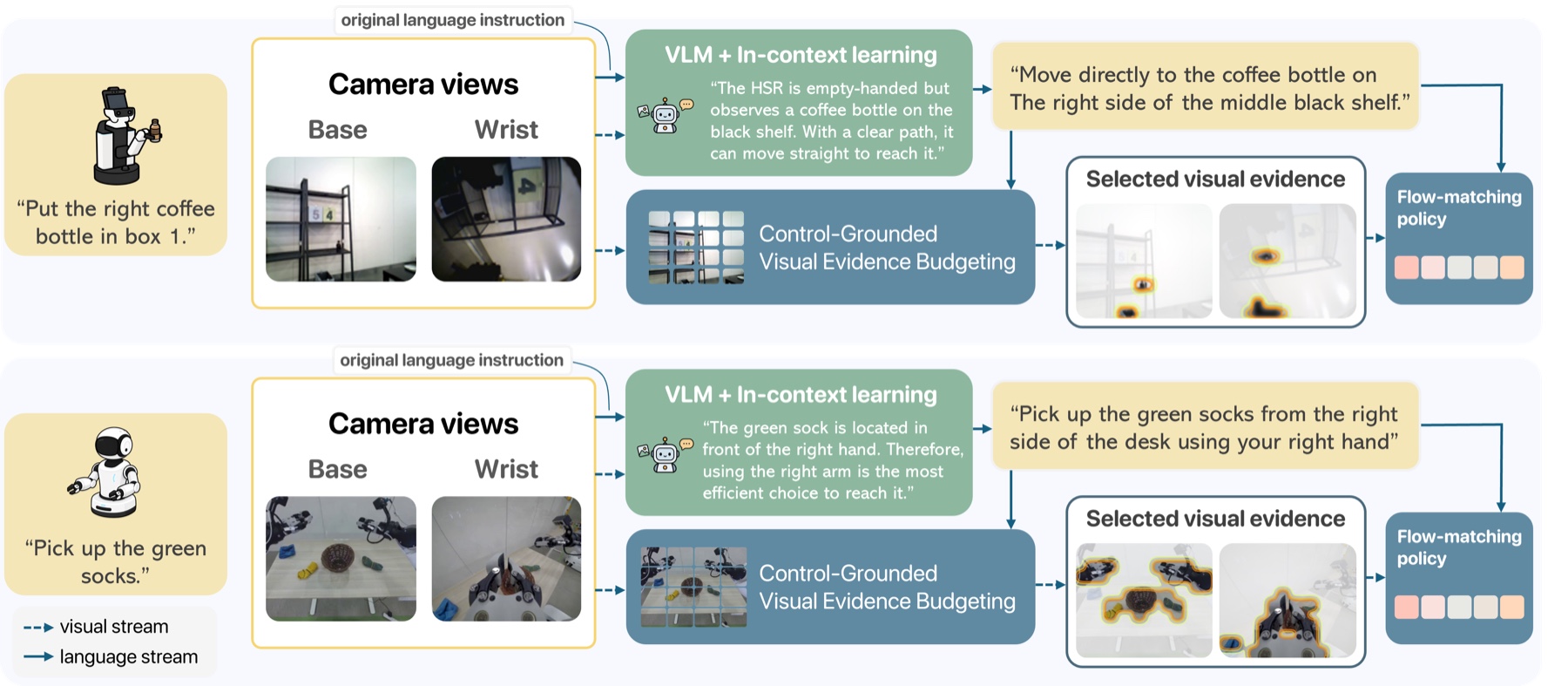}
\end{center}
\vspace{-0.6em}
{\small\color{TextMuted}%
  Generalization suffers when the executor must infer both \emph{what to do now}
  and \emph{what to look at} from a coarse goal alone. S2 resolves this ambiguity
  from both sides: an off-the-shelf VLM rewrites the same high-level instruction
  into state-specific local guidance, while a learned control-grounded visual
  evidence budget shifts the retained evidence from the manipulated object to the
  destination and the relevant local scene context.\par}

\vspace{0.3em}
{\small\sffamily\color{TextMuted}\textbf{Keywords:}~Vision-language-action models, Generalization\par}

\input{sections/01_introduction}
\input{sections/02_related_work}
\input{sections/03_method}
\input{sections/04_experiments}
\input{sections/06_conclusion}

\separator

\subsection*{Acknowledgments}
{\small\color{TextMuted}%
This paper is based on results obtained from a project, JPNP25015, commissioned
by the New Energy and Industrial Technology Development Organization (NEDO). We
thank Ryosuke Takanami and Haru Kondoh for their assistance in the real-robot
experiments.}

\bibliographystyle{corlabbrvnat}
\bibliography{references}

\clearpage
\input{sections/99_appendix}

\end{document}

%% file: sections/01_introduction.tex
\section{Introduction}

Vision-language-action (VLA) models have recently shown strong promise for
robot manipulation, but their generalization remains brittle under distractors,
appearance shifts, and semantically similar tasks that require different
execution strategies \citep{brohan2023rt1, kim2024openvla, black2024pi0}.
General-purpose VLA control is substantially harder than vision-language
understanding alone: the model must not only interpret open-ended goals and
scenes, but also choose appropriate local behaviors and output precise
low-level continuous actions under distractors, appearance shift,
embodiment-specific noise, and tight inference constraints. Expecting a single
low-level policy to absorb all of these demands from finite robot data is
therefore a poor scaling strategy.

Some of this burden is better handled at high level. Modern vision-language
models (VLMs) are stronger at open-ended instruction interpretation and
in-context disambiguation than practical low-level executors, so they can
provide trajectory-aware guidance about which local mode, route, or arm should
be used without forcing the VLA to infer this from a coarse goal alone. Some
burden should also be removed from perception. Not every pixel contributes to
the current control decision, and training an executor to rely on
unconstrained visual context encourages dependence on correlations that may not
generalize. This suggests a cleaner design: rather than forcing one model to
be both planner and executor, we should explicitly disentangle high-level
planning from low-level action generation, and train the executor under a
language-and-vision interface that reduces what it must infer on its own
\citep{ahn2022saycan, myers2025palo, shi2025hirobot}.

In current robot datasets, this burden mismatch appears as avoidable ambiguity
and apparent action multimodality. The same instruction may admit multiple
valid trajectories: an object can be approached from different sides, grasped
at different contact regions, or transported along different collision-free
routes. Some of this variation is irreducible, arising from embodiment,
contact dynamics, or control noise. But some of it is avoidable. Coarse
instructions can alias distinct local behaviors even when a demonstration
already commits to one of them, and full images can expose the policy to
clutter, distractors, and appearance cues that are irrelevant to the current
subtask. Both enlarge the set of modes the low-level policy must explain. We
therefore ask whether executor generalization can be improved by shrinking this
learning problem from both sides at once: providing more informative local
language, and restricting control to a smaller task-sufficient subset of
visual evidence.

We present \stwoname{} (\seelessname{}, \specifymorename{}), a framework built around a more
specific claim: in modular planner-executor VLAs, executor generalization
depends critically on the conditioning interface under which the executor is
trained. Our goal is therefore not merely to add richer language or a learned
visual selection mechanism in isolation, but to train the executor under a
goal-preserving local language interface together with explicit visual evidence
budgeting. \specifymorename{} realizes the language side of this interface. The
original instruction is retained as a stable high-level goal, while refined
trajectory-level and subtask-level descriptions make the demonstrated local
behavior explicit. This separation preserves task identity while reducing
ambiguity about which valid execution mode should be followed. \seelessname{}
realizes the perceptual side. Unlike native attention, which can remain dense
and unconstrained, it introduces an explicit visual evidence budget that trains
the executor to act from a smaller task-sufficient subset of the image rather
than broad visual context. This budget is learned without manual labels,
region annotations, or external VLM supervision: the executor discovers from
the control objective itself which evidence predicts success under the current
subtask. Useful evidence need not correspond to an entire object; it may
instead be a contact surface, target region, clearance cue, or local spatial
relation. In finite-data robot learning, this bottleneck is introduced not for
efficiency but to reduce nuisance dependence and improve generalization
\citep{ryoo2021tokenlearner, cheng2026vlaiap}.

S2 is designed for a modular planner-executor pipeline. At training time, we
relabel demonstrations into refined instructions, ordered subtasks, and
temporal alignments, then fine-tune the VLA under this executor-conditioning
scheme. At deployment, an off-the-shelf VLM provides local subtask guidance
through in-context learning while the low-level VLA focuses on execution.
Because the planner communicates in the same language interface that the
executor is trained to follow, different high-level modules can be swapped
without changing the executor's training target. We therefore do not introduce
a new planner; instead, we study a train-test-consistent executor setup in
which high-level guidance narrows the decision space and explicit evidence
budgeting limits what the policy must rely on perceptually. Across our main
evaluation settings, the results support the same picture: coarse instructions
induce avoidable supervision aliasing, goal-preserving local guidance
outperforms instruction replacement in our main ablations, and explicit
evidence budgeting provides
complementary robustness gains by reducing dependence on broad visual context.
Across eight real-robot tasks on TX-G2 and HSR, S2 raises
mean subtask success from 54.2\% to 79.0\% over $\pi_{0.5}$. Together, these
results suggest that better VLA generalization comes from shrinking the
executor's learning problem in both language and perception, so it can focus
on following the intended behavior rather than recovering it from weak
supervision.

Concretely, we recast generalization in modular planner-executor VLAs as an
executor-conditioning problem, instantiate this view with goal-preserving
hierarchical relabeling and annotation-free control-grounded visual evidence
budgeting, and show empirically that this cleaner interface improves
generalization while remaining compatible with swappable off-the-shelf VLM
planners.

%% file: sections/02_related_work.tex
\section{Related Work}

\paragraph{VLA Generalization and Hierarchical Control.}
Recent VLA models have substantially expanded language-conditioned robot
learning \citep{jang2021bcz, brohan2023rt1, ghosh2024octo, kim2024openvla, black2024pi0},
including compact or cross-embodiment variants \citep{shukor2025smolvla, wang2025bitvla, zheng2025xvla}, yet
benchmarks such as LIBERO and CALVIN still expose brittle generalization under
distribution shift \citep{liu2023libero, mees2022calvin}. In parallel,
language-model-based planners and hierarchical VLA systems decompose high-level
decision making from low-level control
\citep{ahn2022saycan, liang2022codeaspolicies, huang2022inner_monologue, myers2025palo, shi2025hirobot, li2025hamster, long2026vista}.
These works motivate planner-executor decomposition; our focus is the executor
side of that split, and in particular the conditioning interface under which
the low-level policy is trained.

\paragraph{Language Interfaces.}
Several recent works improve low-level controllability through richer command
interfaces. STEER uses dense language grounding for flexible manipulation
\citep{smith2024steer}, Steerable Policies trains VLAs to follow subtasks,
motions, traces, and grounded coordinates \citep{chen2026steerable}, and
ReSteer studies steerability through data refinement \citep{chen2026resteer}.
Pretrained vision-language models have also been used to enrich robot
supervision with more informative language: DIAL and OCI add visual and
object-centric detail \citep{xiao2023dial, wen2024oci}, while NILS and TREAD
scale automatic relabeling to long-horizon trajectories and semantic subtasks
\citep{blank2024nils, kuramshin2025tread}. Our work is closest to this
relabeling line, but targets \emph{specification-induced ambiguity}: the
trajectory already commits to one execution mode while the task instruction
does not. This motivates a goal-preserving local interface that reduces
executor ambiguity without introducing planner-specific control codes. Related
approaches instead strengthen monolithic VLA reasoning more directly through
reasoning-augmented architectures, latent planner-to-policy interfaces, or RL
post-training \citep{zhao2025cotvla, lee2025molmoact, yin2025deepthinkvla, huang2025thinkact, chen2025tgrpo}.

\paragraph{Selective Perception.}
Another relevant line of work studies how policies should focus on task-
relevant perceptual evidence. In robotics, attention-based and object-centric
methods improve robustness by emphasizing relevant visual regions
\citep{abolghasemi2018payattention, devin2017objectcentric, shridhar2022cliport, shridhar2022peract},
while recent work argues for suppressing irrelevant factors or exploiting
locality in visuomotor learning \citep{zhao2024eagle, zhang2025locality, chapin2026spotlighting}.
In parallel, the vision literature and recent VLA work use token selection and
token pruning to reduce redundancy or improve efficiency
\citep{ryoo2021tokenlearner, rao2021dynamicvit, liang2022evit, bolya2023tome, cheng2026vlaiap, li2026bfapp}.
Our visual evidence budgeting instead introduces an explicit, control-grounded
evidence bottleneck: unlike native attention, which can remain dense and
unconstrained, it learns from the control objective which visual evidence is
sufficient for the current behavior, without manual regions, external visual
labels, or efficiency-first objectives.

%% file: sections/03_method.tex
\section{S2: See Less, Specify More}

\stwoname{} defines the executor-conditioning interface in a modular
planner-executor VLA system. \specifymorename{} refines coarse task language
into goal-preserving local supervision, and \seelessname{} learns
annotation-free subtask-conditioned visual evidence restriction directly from
the control objective.
We consider language-conditioned robot demonstrations of the form
$\tau = \big(\{(o_t, a_t)\}_{t=1}^{T}, g\big)$, where $\tau$ denotes a
trajectory of length $T$ paired with an original instruction $g$.
Each observation is $o_t = (I_t^{b}, I_t^{w}, x_t)$, where $I_t^{b}$ and
$I_t^{w}$ are the base and wrist images, respectively, and $x_t$ denotes
proprioceptive state.

Given an episode $\tau$, S2 constructs the hierarchical conditioning interface
$g \rightarrow \tilde{g} \rightarrow \mathcal{S}(\tau)$, where $g$ is the
original instruction, $\tilde{g}$ is a refined trajectory-level instruction,
and
$\mathcal{S}(\tau)=\{(s_i, b_{i-1}, b_i)\}_{i=1}^{N}$ denotes the ordered
subtask sequence, with subtask instruction $s_i$ and temporal span
$[b_{i-1}, b_i)$. The boundaries satisfy
$0 = b_0 \leq b_1 \leq \cdots \leq b_N = T$.

During policy learning, the executor is conditioned on the original
instruction $g$, the active subtask instruction $s_i$, and learned visual
evidence masks over the base and wrist views. At deployment, the planner emits
local guidance in the same form as the subtask instructions used during
training, while the visual masks remain executor-side learned mechanisms
rather than planner outputs.

\subsection{Specify More: Goal-Preserving Hierarchical Relabeling}

For each episode, we first construct the language side of the interface by
producing a refined trajectory instruction
$\tilde{g}$ that explains how the observed trajectory solves the original task.
This relabeling is conditioned on the original instruction $g$ together with
representative images from the episode. The goal is not to paraphrase $g$, but
to resolve omitted execution details such as approach direction, contact
strategy, ordering, obstacle avoidance, or placement behavior while remaining
faithful to the original task identity.

We then decompose the trajectory into an ordered sequence of refined subtasks
$s_1, s_2, \dots, s_N$ using a vision-language relabeling procedure over sparse
trajectory context rather than action heuristics alone. Each subtask is
intended to be directly executable by the low-level policy and specific enough
to distinguish different valid execution modes. We then align each subtask to
the trajectory timeline, yielding frame-aligned supervision of the form
$(s_i, b_{i-1}, b_i)$ for the executor.

The key design choice in \specifymorename{} is that we do not replace the original
instruction with the relabeled one. Instead, we preserve the original
instruction $g$ as the persistent high-level goal and pair it with refined
local language. This separation matters because the original instruction
contains stable task identity, while the refined local instruction resolves the
current execution mode. In other words, the original instruction tells the
policy \emph{what overall task is being solved}, and the refined subtask
instruction tells it \emph{how the current phase should be executed}. Concrete
training-time and evaluation-time prompt-construction procedures for LIBERO,
CALVIN, and LIBERO-Pro are given in
Appendix~\ref{sec:appendix-policy-refinement}.

\subsection{See Less: Learned Visual Evidence Budgeting}

We next construct the perceptual side of the interface. Even after language
ambiguity is reduced, the executor must still determine which parts of the
visual observation matter for the current behavior. Unlike native attention,
which can remain dense and unconstrained, \seelessname{} imposes an explicit
evidence bottleneck on what the executor may rely on for control. We do not
manually specify these regions, nor do we rely on external visual supervision
from a VLM. Instead, the executor learns directly from the control objective
which evidence is sufficient for successful execution under the current
subtask. Useful evidence may include manipulated objects, targets,
contact-relevant surfaces, and local spatial relations or clearance cues;
importantly, it need not correspond to an entire object. We therefore
introduce task-conditioned visual evidence masks $m_t^{b} \in [0,1]^{P_b}$ and
$m_t^{w} \in [0,1]^{P_w}$ over image patches or visual tokens in the base and
wrist views, respectively.

These masks are produced by lightweight learned gate heads inside the
executor. For each view $v \in \{b,w\}$ and token $p$, the gate head takes the
current image token $z_{t,p}^{v}$ together with a pooled summary of the
goal-preserving language context and predicts a soft keep value
$m_{t,p}^{v} = \sigma\!\left(h_v(z_{t,p}^{v}, q_t) / \tau\right)$, where $q_t$
summarizes the current language context and $h_v$ is a small learned gating
network for that view. In our implementation, $h_v$ is a small MLP over the
image token, the pooled language summary, and their elementwise product. We
then apply a nonzero gate floor via
$\hat{m}_{t,p}^{v} = \gamma + (1-\gamma)m_{t,p}^{v}$ and
$\hat{z}_{t,p}^{v} = \hat{m}_{t,p}^{v} z_{t,p}^{v}$, where $\gamma \in [0,1)$.
Because the gate heads are part of the executor itself, masks are produced
online during every forward pass from the current observation and language
context. Training and inference therefore use the same learned masking
mechanism, without external region proposals, mask or box annotations,
planner-provided visual labels, or any other explicit supervision over which
regions should be kept.

\begin{figure*}[t]
    \centering
    \includegraphics[width=\textwidth]{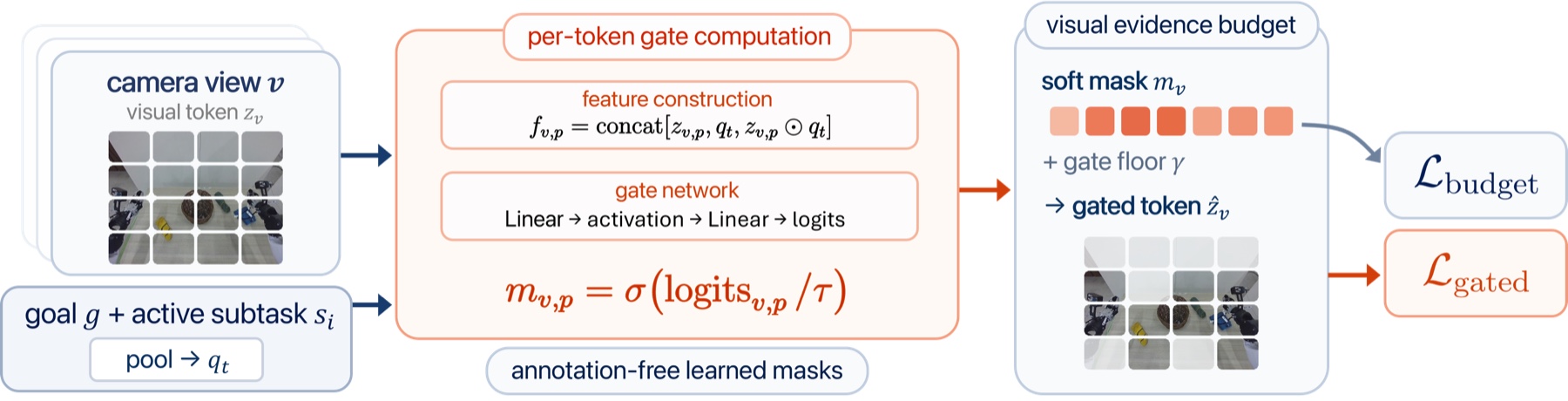}
    \vspace{-0.5em}
    \caption{Control-grounded visual evidence budgeting. For each camera view,
    the executor predicts a soft mask from visual tokens and goal-preserving
    language context, then trains the gated representation with a task loss and
    a budget regularizer, without any region, box, or mask annotation.}
    \label{fig:veb-block}
    \vspace{-0.5em} 
\end{figure*}

To control how much evidence each stream may softly retain, we define separate
visual budgets $\rho_b \in (0,1]$ and $\rho_w \in (0,1]$ for the base and wrist
views. These budgets may differ because the two streams need not carry the
same information: the wrist view often captures contact-local detail, whereas
the base view provides broader spatial context. Their role is not primarily
efficiency, but robustness: the policy is encouraged to solve the current
subtask while depending on less nuisance visual evidence.

To encourage the learned masks to satisfy these targets, we penalize the
deviation between the average pre-floor gate value and the desired soft budget in each
view. Let $\bar{m}_t^{v} = \frac{1}{P_v} \sum_{p=1}^{P_v} m_{t,p}^{v}$ denote
the mean pre-floor gate value in view $v \in \{b,w\}$. We then optimize
\begin{equation}
    \mathcal{L}_{\mathrm{budget}}
    =
    \sum_{v \in \{b,w\}}
    \lambda_{\mathrm{budget}}^{v}
    \mathbb{E}_{\tau,t}\left[\left(\bar{m}_t^{v} - \rho_v\right)^2\right].
\end{equation}
This term constrains how much evidence is softly retained on average while
still allowing the model to decide \emph{which} tokens or patches should be
more strongly retained or suppressed.
The nontrivial selection pressure then comes from maintaining task sufficiency
under the action losses below.

\subsection{Policy Learning}

The executor-conditioning claim of S2 is instantiated through two coupled
paths: an ungated path that preserves the full observation, and a gated path
that applies the learned visual evidence restriction to the image-token
representation under the same language interface. Let
$c_t^{\mathrm{lang}} = (g, s_i)$ denote the language-side conditioning context,
and let $\mathcal{L}_{\mathrm{base}}(\theta; o_t, c_t, a_{t:t+h-1})$ denote a
generic language-conditioned visuomotor training objective, where $o_t$ is the
current observation defined above, $c_t$ is the conditioning context presented
to the policy, $a_{t:t+h-1}$ is the target action chunk, and $h$ is the action
horizon. Here $g$ preserves global task identity and $s_i$ specifies the
current local behavior. The language portion of the conditioning has the same
form as the local guidance emitted by the planner at deployment.

In the current implementation, we retain the ungated base path
\begin{equation}
    \mathcal{L}_{\mathrm{full}}
    =
    \mathcal{L}_{\mathrm{base}}(\theta; o_t, c_t^{\mathrm{lang}}, a_{t:t+h-1}),
\end{equation}
and add a second task loss on the gated token representation
\begin{equation}
    \mathcal{L}_{\mathrm{gated}}
    =
    \mathcal{L}_{\mathrm{base}}(\theta; \tilde{o}_t(m_t^{b}, m_t^{w}), c_t^{\mathrm{lang}}, a_{t:t+h-1}),
\end{equation}
where $\tilde{o}_t(m_t^{b}, m_t^{w})$ denotes the same observation after the
learned token gates are applied to the visual streams. The purpose of
$\mathcal{L}_{\mathrm{gated}}$ is to ensure that the gated representation
remains sufficient for action prediction. Together with
$\mathcal{L}_{\mathrm{budget}}$, it determines \emph{which} evidence can be
discarded: $\mathcal{L}_{\mathrm{budget}}$ constrains how much evidence is
retained, while $\mathcal{L}_{\mathrm{gated}}$ forces the retained subset to
remain control-relevant.

The resulting S2 objective combines the ungated path, the gated path, and the
budget regularizer defined above:
\begin{equation}
    \mathcal{L}_{\mathrm{S2}}
    =
    \mathcal{L}_{\mathrm{full}}
    + \mathcal{L}_{\mathrm{gated}}
    + \mathcal{L}_{\mathrm{budget}}.
\end{equation}
Early gate collapse or trivial all-keep behavior is avoided by three
implementation choices: a nonzero gate floor, the ungated loss
$\mathcal{L}_{\mathrm{full}}$, and an annealed soft-gating schedule that keeps
the gate smoother early in training before sharpening later. The full path
stabilizes optimization, while the coupled
$\mathcal{L}_{\mathrm{gated}} + \mathcal{L}_{\mathrm{budget}}$ objective
learns an annotation-free control-grounded bottleneck rather than a generic
saliency map. Exact backbone-specific instantiations of
$\mathcal{L}_{\mathrm{base}}$, together with the concrete schedules and loss
weights used in our experiments, are given in
Appendix~\ref{sec:appendix-gate-schedules} and
Table~\ref{tab:s2-key-hparams}.

At deployment, the same interface is reused. A user may provide a coarse task
instruction corresponding to $g$, while an off-the-shelf VLM produces refined
local guidance in the same form as the subtask instructions used during
training, via in-context learning rather than task-specific fine-tuning. The
low-level VLA then executes the current detailed instruction under the learned
visual evidence budget. This preserves the intended split of responsibilities:
the VLM resolves which local behavior should be executed, while the low-level
policy focuses on robust execution under reduced visual distraction. Because
the planner communicates through the same guidance form that the executor is
trained to follow, different planners can be swapped in without changing the
executor's training target.

%% file: sections/04_experiments.tex
\section{Experiments}
We evaluate the central executor-conditioning claim of S2. The experiments ask
whether coarse task instructions create conditioning ambiguity, whether the
benefit of refined local language comes from goal-preserving disambiguation
rather than instruction replacement alone, and whether learned visual evidence
restriction provides robustness gains beyond a language-only interface.
\subsection{Experimental Setup}
Unless otherwise specified, all experiments instantiate S2 with a
$\pi_{0.5}$-based executor \citep{physicalintelligence2025pi05} and fine-tune from the released
\texttt{pi05\_base} checkpoint. Unless noted otherwise, we use the
goal-preserving hybrid prompt together with shared VEB at
$\rho_b=\rho_w=0.2$.
Unless otherwise noted, our default relabeling and evaluation-time
prompting/planning configuration uses Kimi K2.5.
Appendix~\ref{sec:appendix-policy-refinement} details the benchmark-specific
prompt refinement, relabeling, and evaluation-time prompting procedures, and
Appendix~\ref{sec:appendix-dataset-composition} summarizes the real-robot
dataset composition.
\subsection{LIBERO-PRO Benchmark}
We first compare the full S2 interface against prior methods across all
LIBERO-PRO perturbation categories.

\begin{table}[t]
    \centering
    \scriptsize
    \setlength{\tabcolsep}{2.7pt}
    \caption{Full LIBERO-PRO perturbation breakdown across the evaluated
    suites. The two S2 rows are lightly highlighted.}
    \label{tab:benchmark-main}
    \renewcommand{\arraystretch}{1.18}
    \adjustbox{max width=\textwidth}{%
    \begin{tabular}{>{\raggedright\arraybackslash}m{4.0cm}|*{4}{>{\centering\arraybackslash}m{0.72cm}}|*{4}{>{\centering\arraybackslash}m{0.72cm}}|*{4}{>{\centering\arraybackslash}m{0.72cm}}|*{4}{>{\centering\arraybackslash}m{0.72cm}}}
        \toprule
        \multirow{2}{4.0cm}{\textbf{Methods}} & \multicolumn{4}{c|}{\texttt{libero\_10}} & \multicolumn{4}{c|}{\texttt{libero\_goal}} & \multicolumn{4}{c|}{\texttt{libero\_object}} & \multicolumn{4}{c}{\texttt{libero\_spatial}} \\
        & Obj. & Pos. & Sem. & Task & Obj. & Pos. & Sem. & Task & Obj. & Pos. & Sem. & Task & Obj. & Pos. & Sem. & Task \\
        \midrule
        OpenVLA-OFT~\citep{kim2025openvlaoft} & 5.5 & 0.0 & 38.5 & 0.0 & 8.5 & 0.0 & 43.5 & 2.5 & 66.5 & 9.5 & 89.0 & 0.0 & 30.0 & 13.5 & 65.0 & 0.0 \\
        X-VLA~\citep{zheng2025xvla} & 61.0 & 1.0 & 70.5 & 19.5 & 71.5 & 1.0 & 94.0 & 7.5 & 91.5 & 4.5 & 98.0 & 0.0 & 89.5 & 0.0 & 69.0 & 38.5 \\
        VLA-Adapter~\citep{wang2025vlaadapter} & 47.0 & 0.0 & 91.0 & 10.0 & 61.0 & 0.0 & 75.0 & 12.0 & 89.0 & 0.0 & 99.0 & 8.0 & 98.0 & 0.0 & \textbf{98.0} & 49.0 \\
        $\pi_{0.5}$~\citepalias{physicalintelligence2025pi05} & 66.0 & 6.0 & 91.0 & 17.0 & 90.0 & 29.0 & 95.0 & 17.0 & 89.0 & 19.0 & 95.0 & 10.0 & \textbf{99.0} & 53.0 & 97.0 & 55.0 \\
        \rowcolor{TableHighlight}
        \mbox{\textbf{S2} (GPT-5.4 nano)} & \textbf{70.0} & 17.0 & 94.0 & 30.5 & \textbf{92.0} & 34.5 & 94.5 & \textbf{44.0} & \textbf{95.5} & \textbf{63.5} & \textbf{100.0} & 24.0 & \textbf{99.0} & 53.5 & 95.0 & \textbf{57.0} \\
        \rowcolor{TableHighlight}
        \mbox{\textbf{S2} (Kimi K2.5)} & 68.0 & \textbf{18.5} & \textbf{94.5} & \textbf{36.0} & 87.0 & \textbf{49.5} & \textbf{97.0} & 43.0 & 90.0 & 62.5 & \textbf{100.0} & \textbf{29.5} & \textbf{99.0} & \textbf{55.0} & 90.5 & 54.5 \\
        \bottomrule
    \end{tabular}
    }
\vspace{-1.2em}
\end{table}

\begin{figure}[!t]
    \centering
    \begin{minipage}[t]{0.53\linewidth}
        \centering
        \includegraphics[width=\linewidth]{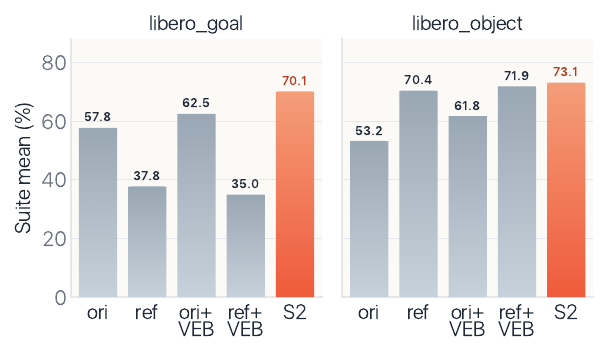}
        \vspace{-1.45em}
        \captionof{figure}{Mean-only object/goal ablation.}
        \label{fig:core-object-goal-means}
    \end{minipage}\hfill
    \begin{minipage}[t]{0.43\linewidth}
        \centering
        \includegraphics[width=\linewidth]{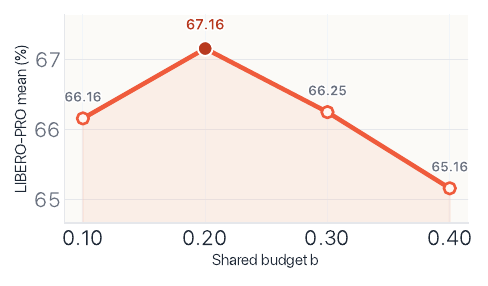}
        \vspace{-1.45em}
        \captionof{figure}{LIBERO-PRO mean vs.\ shared visual evidence budget
        $b$.}
        \label{fig:budget-sweep}
    \end{minipage}
    \vspace{-1.3em}
\end{figure}

The benchmark comparison also highlights that S2 is not tied to a
planner-specific VLM: both S2 rows use the same trained low-level VLA and
differ only in the off-the-shelf planner used to issue refined local guidance:
an open-source Kimi K2.5 planner \citep{kimiteam2026kimi25} and a proprietary
GPT-5.4 nano planner \citep{openai2026gpt54nano}. Across both choices, the
two S2 planner instantiations remain among the strongest on the perturbation
axes that most directly stress executor generalization, with the clearest gains
on \texttt{libero\_goal} and \texttt{libero\_object}. The planner choice
changes which axes peak, but the overall pattern is consistent: the cleaner
executor interface helps the policy identify the intended local behavior while
reducing dependence on task-irrelevant visual factors.
\subsection{LIBERO-PRO Object/Goal Ablation}
The clearest test of the language-side claim is the object/goal ablation on
\texttt{libero\_goal} and \texttt{libero\_object}, which most directly expose
the difference between preserving task identity and merely replacing the
original instruction with richer local text. Figure~\ref{fig:core-object-goal-means}
summarizes five representative settings, and Appendix
Table~\ref{tab:core-object-goal-full} gives the full perturbation breakdown,
including the goal-preserving hybrid without learned VEB.

The result is not simply that more language helps. Refined-only supervision
remains weak on \texttt{libero\_goal}, so local text alone does not solve the
task-identity problem; conversely, recovering task identity without learned VEB
still underperforms the full interface. Only the full S2 interface remains
strong on both suites, supporting the claim that goal-preserving local guidance
and learned visual evidence restriction are both needed.
\subsection{Budget Sensitivity}
Figure~\ref{fig:budget-sweep} shows that the best LIBERO-PRO mean occurs at
$b=0.20$, with nearby budgets remaining competitive. We therefore use
$\rho=0.2$ as the main-paper default.
\subsection{Real-Robot Results}
\begin{wrapfigure}{r}{0.42\linewidth}
    \vspace{-0.8em}
    \centering
    \includegraphics[width=\linewidth]{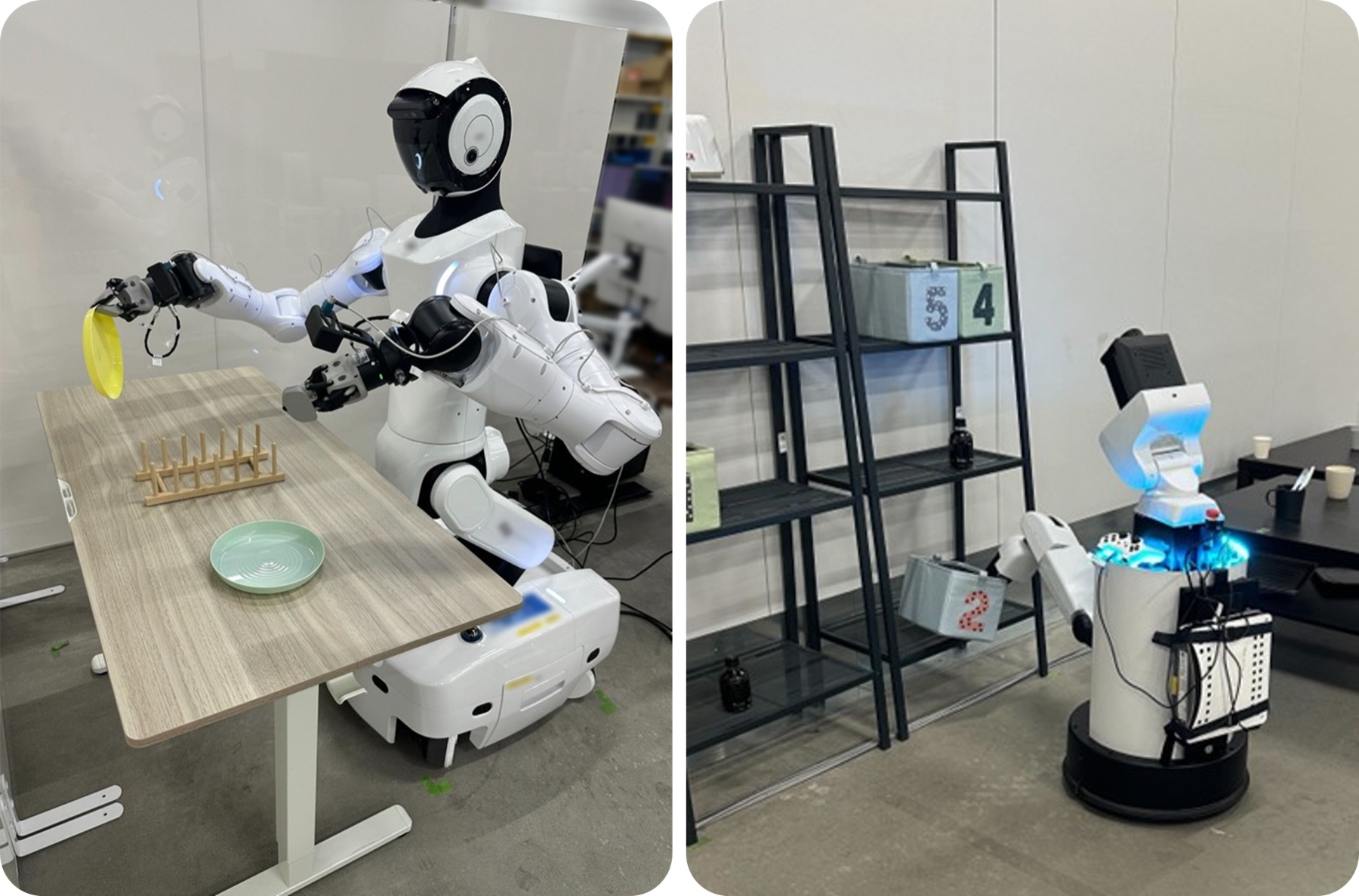}
    \vspace{-1.05em}
    \caption{TX-G2 and HSR.}
    \label{fig:real-robot-scenes}
    \vspace{-1.2em}
\end{wrapfigure}

We additionally report real-robot subtask success on TX-G2 and
HSR. The main text reports mean \emph{Subtask} success, while the full per-task
\emph{Subtask}/\emph{E2E} breakdowns are deferred to
Appendix~\ref{sec:appendix-real-robot}. For fairness, every compared method is
fine-tuned for 150k steps on the corresponding robot dataset and evaluated on
the same RTX 5070 setup; under this constraint, OpenVLA-OFT exceeds available
memory and is therefore omitted from the real-robot comparison.

On TX-G2, each task is evaluated ten times with shared initial states and
object placements across methods (8 near-ID and 2 OOD placements). Because
TX-G2 is bimanual, the policy must also infer which arm should execute the
current behavior. HSR uses six trials per task, adds locomotion, and is
queried at 10\,Hz on TX-G2 and 2\,Hz on HSR. Appendix~\ref{sec:appendix-real-robot-tasks} provides
the detailed task definitions, and Appendix~\ref{sec:appendix-cluttered}
reports TX-G2 cluttered-scene stress tests with unseen distractors and online
object/basket perturbations.

\begin{table}[t]
\centering
\scriptsize
\setlength{\tabcolsep}{3.0pt}
\renewcommand{\arraystretch}{1.16}
\caption{Real-robot mean subtask success on TX-G2 and HSR, with standard
deviation in gray. Full per-task Subtask/E2E tables are deferred to
Appendix~\ref{sec:appendix-real-robot}.}
\vspace{-0.5em}
\label{tab:real-robot-subtask}
\adjustbox{width=0.75\textwidth}{%
\begin{tabular}{>{\raggedright\arraybackslash}m{1.65cm}|*{4}{>{\centering\arraybackslash}m{1.02cm}}|*{4}{>{\centering\arraybackslash}m{1.00cm}}}
\toprule
\multirow[c]{2}{1.65cm}{\textbf{Method}} & \multicolumn{4}{c|}{\textbf{TX-G2}} & \multicolumn{4}{c}{\textbf{HSR}} \\
& \textbf{Cutlery} & \textbf{Bowl} & \textbf{Clothes} & \textbf{Dish} & \textbf{Coffee} & \textbf{Bottles} & \textbf{Box} & \textbf{Mug} \\
\midrule
\mbox{X-VLA} & \stdcell{0.0}{0.0} & \stdcell{7.5}{4.0} & \stdcell{5.0}{2.6} & \stdcell{5.0}{3.4} & \stdcell{8.3}{7.6} & \stdcell{4.2}{3.8} & \stdcell{0.0}{0.0} & \stdcell{25.0}{12.3} \\
\mbox{VLA-Adapter} & \stdcell{0.0}{0.0} & \stdcell{0.0}{0.0} & \stdcell{0.0}{0.0} & \stdcell{0.0}{0.0} & \stdcell{0.0}{0.0} & \stdcell{0.0}{0.0} & \stdcell{0.0}{0.0} & \stdcell{0.0}{0.0} \\
$\pi_{0.5}$ & \stdcell{5.0}{3.2} & \stdcell{47.5}{6.8} & \stdcell{83.3}{4.8} & \stdcell{60.0}{7.2} & \stdcell{83.3}{9.6} & \stdcell{45.8}{8.5} & \stdcell{41.7}{14.0} & \stdcell{66.7}{9.6} \\
\rowcolor{TableHighlight}
\textbf{S2} & \stdcell{\textbf{15.0}}{5.4} & \stdcell{\textbf{67.5}}{6.8} & \stdcell{\textbf{96.7}}{2.2} & \stdcell{\textbf{90.0}}{4.6} & \stdcell{\textbf{100.0}}{0.0} & \stdcell{\textbf{87.5}}{6.6} & \stdcell{\textbf{91.7}}{7.6} & \stdcell{\textbf{83.3}}{9.6} \\
\bottomrule
\end{tabular}%
}
\vspace{-0.5em}
\end{table}

Across both TX-G2 and HSR, S2 improves over $\pi_{0.5}$ on all reported
tasks, with especially large gains on HSR tasks that require combining
locomotion with precise manipulation. VLA-Adapter remains at 0.0 mean
subtask success on both robots despite receiving the same 150k fine-tuning
budget; Appendix~\ref{sec:appendix-real-robot-analysis} discusses diagnostic
evidence and likely causes. These results are consistent with the intended role
of S2: a clearer executor interface improves local subtask execution even when
the robot must combine fine manipulation with arm selection, locomotion, or
both.

\begin{figure*}[!t]
    \centering
    \includegraphics[width=\textwidth]{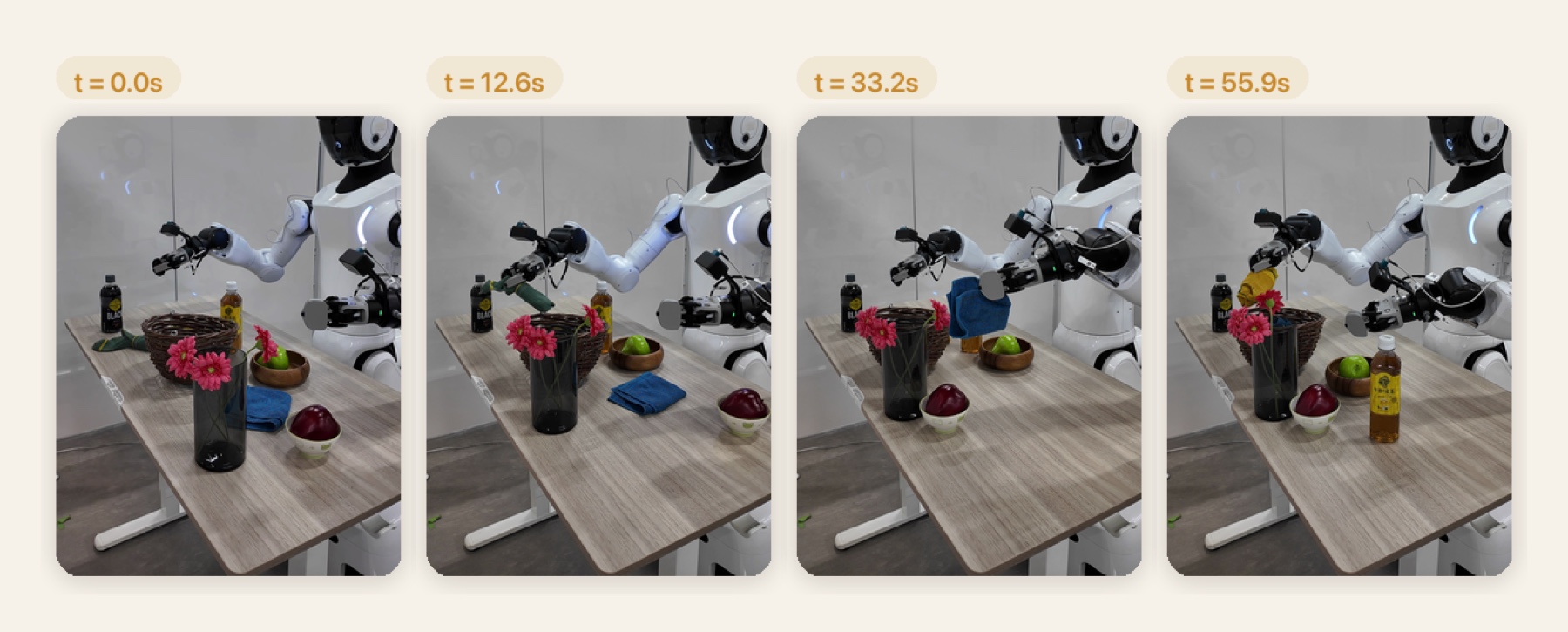}
    \caption{A successful TX-G2 cluttered-scene rollout for the ordered task
    \texttt{green socks $\rightarrow$ handkerchief $\rightarrow$ yellow socks}.
    Most objects on the table (flowers, fruit, bottles) are unseen distractors,
    and a person perturbs object and basket positions during execution, yet S2
    re-grounds the current target online and preserves the required completion
    order. Further examples are given in Appendix~\ref{sec:appendix-cluttered}.}
    \label{fig:cluttered-main}
\end{figure*}

\paragraph{Robustness under unseen clutter.}
Beyond the standard placements, we stress-test S2 on a deliberately cluttered
TX-G2 clothes-sorting setup in which most tabletop objects are unseen distractors
and a person perturbs the scene online during execution.
Figure~\ref{fig:cluttered-main} shows one such rollout: S2 keeps acting on the
instructed garment and basket while ignoring the surrounding clutter, and
completes the ordered sequence despite these perturbations. Additional cluttered
rollouts and their learned masks are reported in
Appendix~\ref{sec:appendix-cluttered}.

\begin{figure*}[!t]
    \centering
    \includegraphics[width=\textwidth]{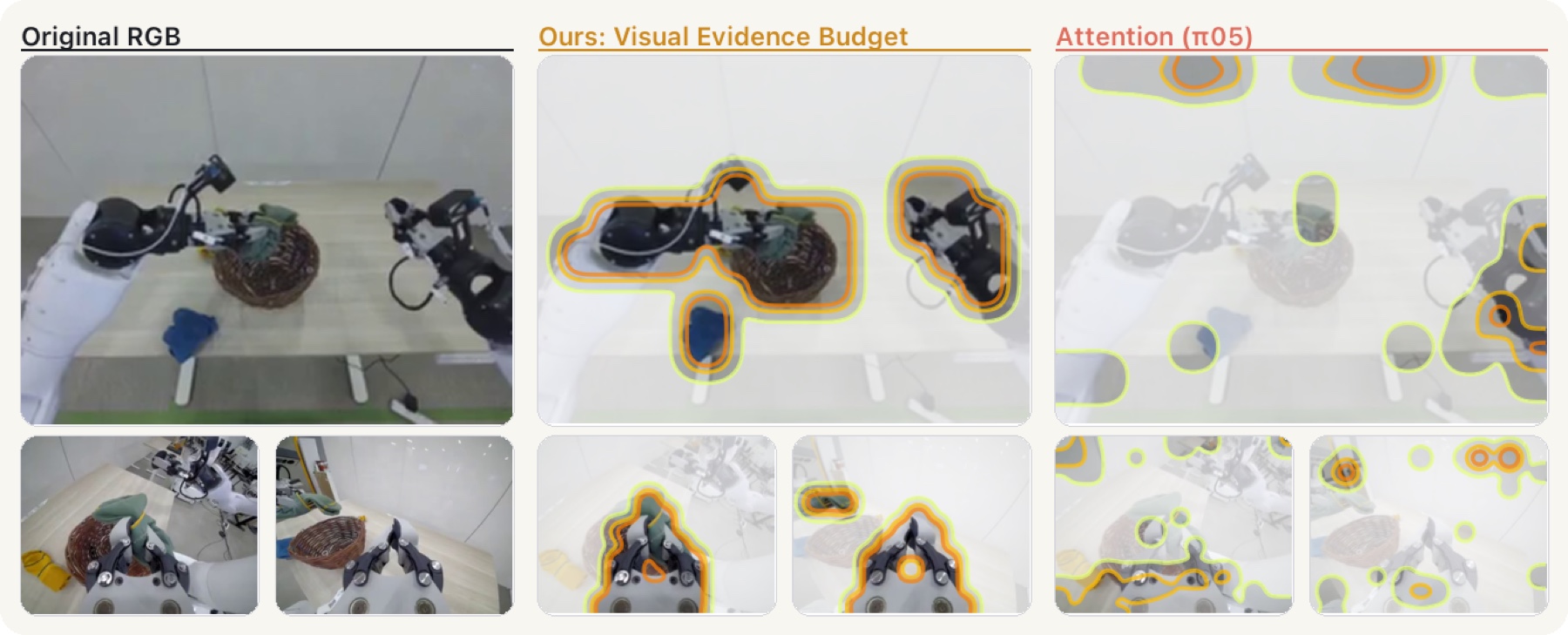}
    \vspace{-1.0em}
    \caption{Qualitative comparison between the learned visual evidence budget
    and the backbone's native attention on representative real-robot
    observations. VEB concentrates on behavior-relevant objects, contact
    regions, and local context, while the backbone attention is more diffuse.
    The learned masks require no region or mask annotation.}
    \label{fig:attention-comparison}
    \vspace{-1.0em}
\end{figure*}

\paragraph{Qualitative evidence selection.}
Figure~\ref{fig:attention-comparison} compares the learned visual evidence
budget with the backbone's native attention on representative real-robot
observations. VEB is more concentrated on the behavior-relevant object,
end-effector, and destination context, while the backbone attention remains
more diffuse and often allocates mass to broad or weakly task-related regions.
The masks are learned without any region or mask annotation; Appendix~\ref{sec:appendix-mask-attention}
provides additional TX-G2 and HSR examples across all real-robot task families.

%% file: sections/06_conclusion.tex
\section{Conclusion}
\label{sec:conclusion}

We presented S2, a framework for improving VLA generalization by training the
executor under a cleaner conditioning interface. Across our evaluated
settings, goal-preserving local guidance outperforms instruction replacement,
and learned visual evidence restriction provides complementary robustness
gains. More broadly, the results suggest that generalizable robot control
benefits from giving the executor a clearer and more task-faithful interface,
rather than forcing it to infer both the intended behavior and the relevant
evidence from weak supervision alone.

%% file: sections/99_appendix.tex
\appendix

\section{Additional Method Details}

\subsection{Backbone-Agnostic Formulation and \texorpdfstring{$\pi_{0.5}$}{pi0.5} Instantiation}

The S2 formulation is largely orthogonal to the choice of underlying
vision-language-action (VLA) backbone. At a high level, `Specify More' modifies
the language interface presented to the policy, and `See Less' modifies how
visual evidence is weighted or suppressed. Neither idea requires a specific
action parameterization. In principle, the same formulation could be combined
with other language-conditioned visuomotor policies, provided they support
conditioning on refined language inputs and some mechanism for task-conditioned
visual gating or masking.

In this work, however, we instantiate S2 using the $\pi_{0.5}$ backbone from
OpenPI \citep{physicalintelligence2025pi05} and fine-tune from the released
`pi05\_base' checkpoint. In our current
setup, the corresponding LIBERO training configuration is `pi05\_libero', which
uses action horizon $10$ and initializes from `pi05\_base'. We use this
instantiation to evaluate the effect of S2, but do not claim that the method
itself is specific to $\pi_{0.5}$.

\subsection{Base Flow-Matching Objective}

In the $\pi_{0.5}$ instantiation, the policy predicts an action chunk using a
flow-matching objective. Let $a$ denote the target action chunk, let
$\epsilon \sim \mathcal{N}(0, I)$ denote Gaussian noise, and let $t \sim p(t)$
denote the flow-matching time variable. The noised action state is
\begin{equation}
    x_t = t\epsilon + (1-t)a,
\end{equation}
and the target flow is
\begin{equation}
    u_t = \epsilon - a.
\end{equation}
Given an observation $o_t$ and a language input $\ell_t$, the backbone predicts
a velocity field $v_{\theta}(x_t, o_t, \ell_t)$ and is trained with the
standard mean-squared flow-matching loss
\begin{equation}
    \mathcal{L}_{\mathrm{base}}
    =
    \mathbb{E}_{(\tau, t, \epsilon)}
    \left[
        \left\|v_{\theta}(x_t, o_t, \ell_t) - u_t \right\|_2^2
    \right].
\end{equation}

\subsection{S2-Conditioned Objective}

Under S2, the generic language input $\ell_t$ is replaced with
goal-preserving hierarchical language context: the original instruction $g$
provides stable task identity, and the active refined subtask instruction
$s_i$ specifies the current execution mode. In addition, `See Less' modifies
the visual context through task-conditioned visual evidence masks
$m_t^{b}, m_t^{w}$. At the level of the backbone objective, this yields
\begin{equation}
    \mathcal{L}_{\mathrm{S2\text{-}cond}}
    =
    \mathbb{E}_{(\tau, t, \epsilon)}
    \left[
        \left\|v_{\theta}(x_t, o_t, g, s_i, m_t^{b}, m_t^{w}) - u_t \right\|_2^2
    \right].
\end{equation}
This formulation emphasizes that S2 changes the information made available to
the policy, while remaining compatible with the underlying backbone objective.

\subsection{Visual Budget Gate Architecture}

The learned visual budget gate operates directly on image-token embeddings
before they are concatenated with prompt tokens into the backbone prefix. For
each image token $z_{t,p}^{v}$, the gate head receives three feature blocks:
\begin{equation}
    \phi_{t,p}^{v}
    =
    \big[
        z_{t,p}^{v},\;
        q_t,\;
        z_{t,p}^{v} \odot q_t
    \big],
\end{equation}
where $q_t$ is a pooled prompt summary and $\odot$ denotes elementwise
product. In the current implementation, each view uses a lightweight two-layer
MLP
\begin{equation}
    \ell_{t,p}^{v} = W_2 \,\mathrm{swish}(W_1 \phi_{t,p}^{v}),
\end{equation}
followed by temperature-scaled sigmoid gating
\begin{equation}
    m_{t,p}^{v} = \sigma(\ell_{t,p}^{v} / \tau).
\end{equation}
The effective multiplicative gate applied to the token embedding is
\begin{equation}
    \hat{m}_{t,p}^{v}
    =
    \gamma + (1-\gamma)m_{t,p}^{v},
    \qquad
    \hat{z}_{t,p}^{v}
    =
    \hat{m}_{t,p}^{v} z_{t,p}^{v},
\end{equation}
where $\gamma$ is the configured gate floor. The floor ensures that every token
retains a small residual contribution even when its learned gate is near zero.

\subsection{Visual Evidence Budget Objective}

In our $\pi_{0.5}$ implementation, visual evidence budgeting is optimized with
both an ungated prediction path and a gated path. The total training objective
takes the form
\begin{equation}
    \mathcal{L}_{\mathrm{S2}}
    =
    \mathcal{L}_{\mathrm{full}}
    + \lambda_{\mathrm{gated}} \mathcal{L}_{\mathrm{gated}}
    + \lambda_{\mathrm{budget}}^{b} \mathcal{L}_{\mathrm{budget}}^{b}
    + \lambda_{\mathrm{budget}}^{w} \mathcal{L}_{\mathrm{budget}}^{w},
\end{equation}
where $\mathcal{L}_{\mathrm{full}}$ is the base flow-matching loss computed on
the ungated representation, $\mathcal{L}_{\mathrm{gated}}$ is the
corresponding loss on the gated representation,
$\mathcal{L}_{\mathrm{budget}}^{b}$ and $\mathcal{L}_{\mathrm{budget}}^{w}$
penalize deviations from the target base and wrist soft keep ratios.
Concretely, if
\begin{equation}
    \bar{m}_t^{b} = \frac{1}{P_b} \sum_{p=1}^{P_b} m_{t,p}^{b},
    \qquad
    \bar{m}_t^{w} = \frac{1}{P_w} \sum_{p=1}^{P_w} m_{t,p}^{w},
\end{equation}
then the budget penalties take the form
\begin{equation}
    \mathcal{L}_{\mathrm{budget}}^{b}
    =
    \mathbb{E}_{\tau,t}\left[\left(\bar{m}_t^{b} - \rho_b\right)^2\right],
    \qquad
    \mathcal{L}_{\mathrm{budget}}^{w}
    =
    \mathbb{E}_{\tau,t}\left[\left(\bar{m}_t^{w} - \rho_w\right)^2\right].
\end{equation}
This matches the implementation choice of penalizing the squared error between
the per-sample mean pre-floor gate value and the desired soft keep ratio in
each view.

This objective is specific to our $\pi_{0.5}$ implementation of `See Less'. The
broader S2 idea does not depend on this exact loss decomposition; what is
essential is that the policy is trained under refined hierarchical language
conditioning and task-conditioned suppression of irrelevant visual evidence.

\subsection{Gated Loss}

Beyond the base task loss and the budget penalties, our implementation also
uses a gated prediction loss. The gated loss has the same form as the base
task loss, but is computed after applying the learned
visual evidence masks to the image-token representation:
\begin{equation}
    \mathcal{L}_{\mathrm{gated}}
    =
    \mathbb{E}_{(\tau, t, \epsilon)}
    \left[
        \left\|v_{\theta}(x_t, o_t, g, s_i, m_t^{b}, m_t^{w}) - u_t \right\|_2^2
    \right].
\end{equation}
In other words, the model is trained both on the original ungated visual
representation and on the gated representation. This is important because the
budget loss alone only constrains the average pre-floor gate value; without
$\mathcal{L}_{\mathrm{gated}}$, many degenerate keep patterns would satisfy the
budget equally well. The gated task loss breaks this degeneracy by penalizing
any gate pattern that removes evidence needed for action prediction.

\subsection{Schedules and Early-Training Stabilization}
\label{sec:appendix-gate-schedules}

Several scheduled quantities are used to keep the gate stable early in
training. The gate sigmoid temperature (denoted here as
$\tau_{\mathrm{gate}}$ to distinguish it from the trajectory variable $\tau$)
is annealed linearly from $\tau_{\mathrm{start}} = 1.0$ to
$\tau_{\mathrm{end}} = 0.7$:
\begin{equation}
    \tau_{\mathrm{gate}}(k)
    =
    \tau_{\mathrm{start}}
    +
    \left(\tau_{\mathrm{end}} - \tau_{\mathrm{start}}\right)
    \frac{k}{K-1},
\end{equation}
where $k$ is the current optimization step and $K$ is the total number of
training steps. Equivalently, early training uses a higher temperature, so the
gate remains smoother and less saturated, while later training uses a lower
temperature, producing sharper keep/suppress decisions that better match the
target soft evidence budget. At inference time, we use the sharpened endpoint value
$\tau_{\mathrm{end}}$. Combined with the ungated base path and the nonzero gate
floor, this schedule is the main mechanism used to avoid immediate gate
collapse while still converging to selective evidence restriction.

\subsection{Per-View Visual Budgets}

Our formulation permits separate visual evidence budgets for the base and wrist
views,
\begin{equation}
    \rho_b \in (0,1], \qquad \rho_w \in (0,1].
\end{equation}
This is important because the two views need not contain the same type or
amount of useful information. In particular, the wrist view often contains more
contact-local detail, while the base view provides broader scene context and
may contain different forms of clutter or distractors.

In the main experiments, however, we set $\rho_b = \rho_w = 0.2$ in order to
reduce the experimental search space and isolate the effect of visual evidence
budgeting itself without introducing an additional per-view hyperparameter
sweep. This shared-budget setting should be understood as an experimental
simplification rather than a limitation of the method.

\subsection{Key Hyperparameters}

Table~\ref{tab:s2-key-hparams} summarizes the most important gate-related and
training-level hyperparameters used in our experiments. The gate-specific
values are shared across our strongest TX-G2 and HSR runs, while LIBERO differs
mainly in the soft keep-target sweep and some training-level schedule choices.

\begin{table}[h]
    \centering
    \scriptsize
    \setlength{\tabcolsep}{4pt}
    \caption{Key S2 hyperparameter values used in the main experiment families.}
    \label{tab:s2-key-hparams}
    \renewcommand{\arraystretch}{1.12}
    \begin{tabular}{>{\raggedright\arraybackslash}m{5.0cm} >{\centering\arraybackslash}m{2.4cm}}
        \toprule
        Hyperparameter & Typical value(s) \\
        \midrule
        Gate-head hidden dim & 256 \\
        Gate floor $\gamma$ & 0.1 \\
        Soft keep targets $\rho_b, \rho_w$ & 0.2 \\
        $\lambda_{\mathrm{gated}}$ & 1.0 \\
        \shortstack[l]{$\lambda_{\mathrm{budget}}^{b},$\\$\lambda_{\mathrm{budget}}^{w}$} &
        0.1 / 0.1 \\
        \shortstack[l]{$\tau_{\mathrm{start}},$\\$\tau_{\mathrm{end}}$} &
        1.0 $\rightarrow$ 0.7 \\
        Batch size & \shortstack[c]{LIBERO: 128\\TX-G2/HSR: 64} \\
        \bottomrule
    \end{tabular}
\end{table}

\subsection{Subtask Alignment Noise and Planner Sensitivity}

The current training path uses hard frame-aligned subtask spans. As a result,
subtask-boundary errors appear directly as supervision noise: a frame near a
transition may be paired with a subtask instruction that is slightly too early
or too late. We mitigate this in practice by using trajectory-aware relabeling
and semantically coarse subtask phases, but we do not explicitly model
boundary uncertainty in the current policy-training path.

At deployment, the executor assumes that the current subtask instruction is
approximately correct. If the planner emits an incorrect subtask or revises it
poorly, the executor will typically attempt to follow that incorrect local
guidance rather than recover autonomously. This is therefore a current failure
mode of the modular closed-loop setup: planner errors appear to the executor as
conditioning errors.

\section{Additional Experimental Details}

\subsection{Policy-Facing Language Refinement}
\label{sec:appendix-policy-refinement}

This subsection describes how we construct policy-facing language prompts at
training time and, when applicable, how we adapt them at evaluation time. The
procedure is benchmark-specific: LIBERO requires phase-aware relabeling from
demonstrations, CALVIN exposes an official current goal or phase signal during
evaluation, and LIBERO-Pro adds a low-frequency closed-loop prompt planner on
top of LIBERO-trained executors.
Unless otherwise specified, the default VLM for both dataset relabeling and
evaluation-time prompting/planning is Kimi K2.5.

\paragraph{Shared trajectory-level refinement.}
We first convert each coarse task annotation into a trajectory-specific
executable instruction. A VLM receives the original task instruction together
with a sparse set of representative keyframes and rewrites the task into one
instruction that describes how the current trajectory solves it. The refiner
may add visually supported execution details such as spatial anchors, approach
direction, grasp style, motion ordering, or placement strategy, but it must
preserve task identity and avoid retrospective commentary. To reduce task
drift, we provide contrastive refined instructions from other demonstrations of
the same task as additional references and run candidate generation, grounded
selection, and QC repair before export. A compact prompt skeleton is:

\begin{quote}
    \scriptsize
    \texttt{TASK: <original instruction>}\\
    \texttt{KEYFRAMES: initial scene + sparse trajectory images}\\
    \texttt{ADDITIONAL REFERENCES: refined instructions from other visibly}\\
    \texttt{\hspace*{1.3em}different strategies of the same task}\\
    \texttt{GOAL: rewrite the task into one imperative instruction for the}\\
    \texttt{\hspace*{1.3em}current trajectory's strategy}\\
    \texttt{OUTPUT: one executable instruction for this trajectory only}\\
    \texttt{STYLE: concrete, scene-grounded, imperative, and concise}\\
    \texttt{CONSTRAINTS: preserve task identity; add only visually supported}\\
    \texttt{\hspace*{1.3em}details; avoid unsupported objects, frame numbers,}\\
    \texttt{\hspace*{1.3em}retrospective narration, and free-form commentary}
\end{quote}

\paragraph{LIBERO training-time prompts.}
For LIBERO, the trajectory-level instruction serves as an intermediate
representation for phase-aware supervision. We decompose each trajectory into
an ordered sequence over a small canonical manipulation vocabulary:
\texttt{approach}, \texttt{engage}, \texttt{execute},
\texttt{disengage}, and \texttt{transit}. The decomposition returns
phase-labeled subtasks with coarse image anchors, which are then grounded to
frame spans and locally refined while keeping the phase structure fixed. The
final rendered language view is written into the LeRobot \texttt{task} field,
so the exported prompt surface is part of the training specification. In the
main paper, this view is the goal-preserving hybrid interface consisting of the
original instruction together with the active refined subtask instruction. The
phase-decomposition interface is:

\begin{quote}
    \scriptsize
    \texttt{INPUTS: original task, refined trajectory instruction, ordered}\\
    \texttt{\hspace*{1.3em}keyframes}\\
    \texttt{PHASE VOCAB: approach, engage, execute, disengage, transit}\\
    \texttt{GOAL: return an ordered phase plan grounded to the observed}\\
    \texttt{\hspace*{1.3em}trajectory rather than an abstract language plan}\\
    \texttt{OUTPUT: JSON list with phase, subtask, and image-anchor fields}\\
    \texttt{STYLE: each subtask should be locally executable and phase-aware}\\
    \texttt{CONSTRAINTS: preserve object identity and phase order; refine}\\
    \texttt{\hspace*{1.3em}wording locally, but do not split, reorder, or}\\
    \texttt{\hspace*{1.3em}rename segments once the coarse decomposition is set}
\end{quote}

\paragraph{CALVIN training and inference.}
CALVIN already exposes a current goal or phase signal at evaluation time, so
we do not perform LIBERO-style phase discovery. For training data, we apply
only trajectory-level refinement and render the prompt as
\texttt{High-level goal: <original goal>} together with
\texttt{Current instruction: <refined full-trajectory instruction>}. At
inference time, the refiner receives the official current goal or phase, recent
observations, and robot state, and returns one short executable instruction in
strict JSON form under a single \texttt{current\_instruction} field. In the
goal-change mode used in our experiments, this instruction is recomputed only
when the official CALVIN goal or phase changes, preventing unnecessary
high-frequency language churn while preserving the benchmark-provided task
identity. The inference-time prompt skeleton is:

\begin{quote}
    \scriptsize
    \texttt{OFFICIAL CURRENT GOAL/PHASE: <goal>}\\
    \texttt{RECENT OBSERVATIONS: <images>}\\
    \texttt{ROBOT STATE: <state vector>}\\
    \texttt{GOAL: write one short executable instruction for the official}\\
    \texttt{\hspace*{1.3em}current goal or phase only}\\
    \texttt{OUTPUT: JSON with a single current\_instruction field}\\
    \texttt{STYLE: short, imperative, scene-grounded, and policy-facing}\\
    \texttt{CONSTRAINTS: preserve the official goal exactly; return one short}\\
    \texttt{\hspace*{1.3em}executable instruction; do not expand into a plan,}\\
    \texttt{\hspace*{1.3em}change the goal, or copy retrieved scene details}
\end{quote}

\paragraph{LIBERO-Pro closed-loop prompting.}
LIBERO-Pro reuses LIBERO-trained executors but evaluates them under a stricter
closed-loop prompting protocol. At the start of an episode, an off-the-shelf
VLM compiles a trajectory-level strategy from the original task and current
scene, converts it into an ordered phase plan, and activates the first local
instruction. During execution, the planner periodically reviews recent
observations after a minimum commitment window and chooses among
\texttt{keep\_current}, \texttt{refine\_current},
\texttt{next\_subtask}, \texttt{recover}, and
\texttt{task\_complete}. The planner never outputs robot actions directly; it
only revises the policy-facing prompt, while continuous actions remain the
responsibility of the trained VLA executor. The review interface is:

\begin{quote}
    \scriptsize
    \texttt{INPUTS: original task, current active instruction, recent views,}\\
    \texttt{\hspace*{1.3em}recent robot progress}\\
    \texttt{ALLOWED DECISIONS: keep\_current, refine\_current, next\_subtask,}\\
    \texttt{\hspace*{1.3em}recover, task\_complete}\\
    \texttt{GOAL: revise the active policy prompt conservatively, only when}\\
    \texttt{\hspace*{1.3em}recent evidence indicates progress, completion, or drift}\\
    \texttt{OUTPUT: one decision + optional revised short imperative}\\
    \texttt{\hspace*{1.3em}instruction compatible with the training prompt form}\\
    \texttt{CONSTRAINTS: default to keep\_current when execution remains on}\\
    \texttt{\hspace*{1.3em}track; advance only when the current phase is visibly}\\
    \texttt{\hspace*{1.3em}satisfied; revise language without emitting actions}
\end{quote}

\subsection{Dataset Composition}
\label{sec:appendix-dataset-composition}

This subsection summarizes the current real-robot datasets used in the TX-G2
and HSR experiments.

\paragraph{TX-G2 dataset composition.}
The current TX-G2 dataset contains 1,198 source trajectories and 5,287
primitive-action (PA) segments after segmentation. Table~\ref{tab:g2-dataset-composition}
summarizes the source-trajectory count for each task family, and the canonical
PA sequence for each family is listed below for reference.

\begin{table}[h]
    \centering
    \scriptsize
    \setlength{\tabcolsep}{6pt}
    \caption{TX-G2 dataset composition by task family.}
    \label{tab:g2-dataset-composition}
    \renewcommand{\arraystretch}{1.12}
    \begin{tabular}{lc}
        \toprule
        Task family & Source trajs \\
        \midrule
        Bowl Stacking & 129 \\
        Clothes Sorting & 514 \\
        Cutlery Transfer & 206 \\
        Dish Racking & 349 \\
        \bottomrule
    \end{tabular}
\end{table}

\paragraph{Canonical PA sequences.}
\texttt{Bowl Stacking}: Pick up the yellow bowl from the desk $\rightarrow$
stack the yellow bowl on the grey bowl $\rightarrow$ pick up the light blue
bowl from the desk $\rightarrow$ stack the light blue bowl on the yellow bowl.
\texttt{Clothes Sorting}: Pick up the green socks from the desk $\rightarrow$
place the green socks into the basket $\rightarrow$ pick up the handkerchief
from the desk $\rightarrow$ place the handkerchief into the basket
$\rightarrow$ pick up the yellow socks from the desk $\rightarrow$ place the
yellow socks into the basket. \texttt{Cutlery Transfer}: Pick up the light
blue spoon from the grey bowl $\rightarrow$ place the light blue spoon in the
yellow bowl $\rightarrow$ pick up the pink fork from the grey bowl
$\rightarrow$ place the pink fork in the yellow bowl. \texttt{Dish Racking}:
Pick up the yellow dish from the desk $\rightarrow$ place the yellow dish in
the wooden dish rack $\rightarrow$ pick up the green dish from the desk
$\rightarrow$ place the green dish in the wooden dish rack.

\paragraph{HSR dataset composition.}
The current HSR dataset contains 1,205 source trajectories. Table~\ref{tab:hsr-dataset-composition}
summarizes the source-trajectory count for each task family, and the canonical
PA sequence for each family is listed below for reference.

\begin{table}[h]
    \centering
    \scriptsize
    \setlength{\tabcolsep}{6pt}
    \caption{HSR dataset composition by task family.}
    \label{tab:hsr-dataset-composition}
    \renewcommand{\arraystretch}{1.12}
    \begin{tabular}{lc}
        \toprule
        Task family & Source trajs \\
        \midrule
        Mug Rectangle & 399 \\
        Coffee Bottle $\rightarrow$ Box 1 & 295 \\
        Box Rearrangement & 195 \\
        Coffee Bottles $\rightarrow$ Table & 316 \\
        \bottomrule
    \end{tabular}
\end{table}

\paragraph{Canonical HSR PA sequences.}
\texttt{Mug Rectangle} (instruction: form a rectangle by relocating the mug
that is not at a rectangle corner): pick up the mug that is not at a
rectangle corner $\rightarrow$ place the mug at the missing rectangle corner.
\texttt{Coffee Bottle $\rightarrow$ Box 1} (instruction: place the coffee
bottle on the right into the box labeled \texttt{1}): pick up the coffee
bottle on the right $\rightarrow$ place the coffee bottle into the box labeled
\texttt{1}. \texttt{Box Rearrangement} (instruction: relocate box number
\texttt{2} beside the box numbered \texttt{1}): pick up the box labeled
\texttt{2} $\rightarrow$ place the box next to the box labeled \texttt{1}.
\texttt{Coffee Bottles $\rightarrow$ Table} (instruction: relocate the two
coffee bottles from the shelves to the table beside the mugs): pick up the
right-hand bottle from the shelves $\rightarrow$ place it next to any mug on
the table $\rightarrow$ pick up the left-hand bottle from the shelves
$\rightarrow$ place it next to a mug on the table that does not already have a
bottle.

\subsection{Backbone and Fine-Tuning Setup}

All experiments in the main paper instantiate S2 with the $\pi_{0.5}$
backbone and fine-tune from the `pi05\_base' checkpoint. Unless otherwise
specified, we use the `pi05\_libero' configuration for LIBERO experiments and
the corresponding OpenPI fine-tuning pipeline.

\subsection{Instruction Views}

The relabeling pipeline produces multiple language views, including the
original instruction, refined trajectory instruction, and refined subtask
instruction. The main paper focuses on the goal-preserving hybrid setting,
which conditions the policy on both the original instruction and the active
refined subtask instruction.

\subsection{Visual Budget Defaults}

Our implementation supports separate base and wrist soft keep targets. While
the method is view-specific by design, the main experiments use a shared-budget
setting with $\rho_b=\rho_w=0.2$ unless otherwise noted.

\subsection{CALVIN ABC\texorpdfstring{$\rightarrow$}{->}D Benchmark}

We additionally report CALVIN ABC$\rightarrow$D average sequence length in
Table~\ref{tab:calvin-abcd}. Because CALVIN already evaluates staged
long-horizon tasks with explicit task boundaries, this benchmark does not
exercise closed-loop phase detection in the same way as LIBERO-PRO. We
therefore instantiate S2 only through its Specify More language interface,
without an additional phase detector or phase-change module. The \textbf{S2}
row in Table~\ref{tab:calvin-abcd} refers to this language-side instantiation.
We compare against reported results for OpenVLA-OFT, UniVLA, VLA-Adapter, and X-VLA
\citep{kim2025openvlaoft,wang2025vlaadapter,wang2025univla,zheng2025xvla}.

\begin{table}[h]
    \centering
    \scriptsize
    \setlength{\tabcolsep}{5pt}
    \caption{CALVIN ABC$\rightarrow$D average sequence length. Higher is better.
    Here \textbf{S2} denotes the CALVIN instantiation that uses only the
    Specify More language interface.}
    \label{tab:calvin-abcd}
    \renewcommand{\arraystretch}{1.12}
    \begin{tabular}{llc}
        \toprule
        Suite & Method & Avg. Len. $\uparrow$ \\
        \midrule
        CALVIN ABC$\rightarrow$D & OpenVLA-OFT & 3.27 \\
        CALVIN ABC$\rightarrow$D & UniVLA & 3.80 \\
        CALVIN ABC$\rightarrow$D & VLA-Adapter & 4.42 \\
        CALVIN ABC$\rightarrow$D & X-VLA & \textbf{4.43} \\
        CALVIN ABC$\rightarrow$D & $\pi_{0.5}$ & 3.92 \\
        \rowcolor{TableHighlight}
        CALVIN ABC$\rightarrow$D & \textbf{S2} & 3.95 \\
        \bottomrule
    \end{tabular}
\end{table}

\subsection{Real-Robot End-to-End Results}
\label{sec:appendix-real-robot}

Main-text real-robot comparisons report only mean subtask success. Tables
\ref{tab:real-robot-g2-full} and \ref{tab:real-robot-hsr-full} provide the
full per-task \emph{Subtask}/\emph{E2E} breakdowns, where \emph{E2E} counts an
episode as successful only if every subtask succeeds consecutively from start
to finish. As in the main paper, all compared methods in the real-robot study
are fine-tuned for 150k steps before evaluation. OpenVLA-OFT is omitted from
these tables because it exceeds memory on the shared RTX 5070 evaluation
setup. TX-G2 uses ten trials per task, while HSR uses six.

\begin{table}[h]
    \centering
    \scriptsize
    \setlength{\tabcolsep}{2.4pt}
    \caption{Full TX-G2 real-robot results. For each task, we report mean
    subtask success with standard deviation and end-to-end success.}
    \label{tab:real-robot-g2-full}
    \renewcommand{\arraystretch}{1.16}
    \adjustbox{max width=\textwidth}{%
    \begin{tabular}{>{\raggedright\arraybackslash}m{2.00cm}|*{2}{>{\centering\arraybackslash}m{1.08cm}}|*{2}{>{\centering\arraybackslash}m{1.08cm}}|*{2}{>{\centering\arraybackslash}m{1.08cm}}|*{2}{>{\centering\arraybackslash}m{1.08cm}}}
        \toprule
        \multirow{2}{2.00cm}{\textbf{Method}} & \multicolumn{2}{c|}{\textbf{Cutlery}} & \multicolumn{2}{c|}{\textbf{Bowl}} & \multicolumn{2}{c|}{\textbf{Clothes}} & \multicolumn{2}{c}{\textbf{Dish}} \\
        & Subtask & E2E & Subtask & E2E & Subtask & E2E & Subtask & E2E \\
        \midrule
        X-VLA & \stdcell{0.0}{0.0} & \textbf{0.0} & \stdcell{7.5}{4.0} & 0.0 & \stdcell{5.0}{2.6} & 0.0 & \stdcell{5.0}{3.4} & 0.0 \\
        VLA-Adapter & \stdcell{0.0}{0.0} & \textbf{0.0} & \stdcell{0.0}{0.0} & 0.0 & \stdcell{0.0}{0.0} & 0.0 & \stdcell{0.0}{0.0} & 0.0 \\
        $\pi_{0.5}$ & \stdcell{5.0}{3.2} & \textbf{0.0} & \stdcell{47.5}{6.8} & 0.0 & \stdcell{83.3}{4.8} & 50.0 & \stdcell{60.0}{7.2} & 20.0 \\
        \rowcolor{TableHighlight}
        \textbf{S2} & \stdcell{\textbf{15.0}}{5.4} & \textbf{0.0} & \stdcell{\textbf{67.5}}{6.8} & \textbf{30.0} & \stdcell{\textbf{96.7}}{2.2} & \textbf{90.0} & \stdcell{\textbf{90.0}}{4.6} & \textbf{70.0} \\
        \bottomrule
    \end{tabular}
    }
\end{table}

\begin{table}[h]
    \centering
    \scriptsize
    \setlength{\tabcolsep}{2.4pt}
    \caption{Full HSR real-robot results. For each task, we report mean
    subtask success with standard deviation and end-to-end success.}
    \label{tab:real-robot-hsr-full}
    \renewcommand{\arraystretch}{1.16}
    \adjustbox{max width=\textwidth}{%
    \begin{tabular}{>{\raggedright\arraybackslash}m{2.00cm}|*{2}{>{\centering\arraybackslash}m{1.08cm}}|*{2}{>{\centering\arraybackslash}m{1.08cm}}|*{2}{>{\centering\arraybackslash}m{1.08cm}}|*{2}{>{\centering\arraybackslash}m{1.08cm}}}
        \toprule
        \multirow{2}{2.00cm}{\textbf{Method}} & \multicolumn{2}{c|}{\textbf{Coffee}} & \multicolumn{2}{c|}{\textbf{Bottles}} & \multicolumn{2}{c|}{\textbf{Box}} & \multicolumn{2}{c}{\textbf{Mug}} \\
        & Subtask & E2E & Subtask & E2E & Subtask & E2E & Subtask & E2E \\
        \midrule
        X-VLA & \stdcell{8.3}{7.6} & 0.0 & \stdcell{4.2}{3.8} & 0.0 & \stdcell{0.0}{0.0} & 0.0 & \stdcell{25.0}{12.3} & 16.7 \\
        VLA-Adapter & \stdcell{0.0}{0.0} & 0.0 & \stdcell{0.0}{0.0} & 0.0 & \stdcell{0.0}{0.0} & 0.0 & \stdcell{0.0}{0.0} & 0.0 \\
        $\pi_{0.5}$ & \stdcell{83.3}{9.6} & 66.7 & \stdcell{45.8}{8.5} & 0.0 & \stdcell{41.7}{14.0} & 16.7 & \stdcell{66.7}{9.6} & 33.3 \\
        \rowcolor{TableHighlight}
        \textbf{S2} & \stdcell{\textbf{100.0}}{0.0} & \textbf{100.0} & \stdcell{\textbf{87.5}}{6.6} & \textbf{66.7} & \stdcell{\textbf{91.7}}{7.6} & \textbf{83.3} & \stdcell{\textbf{83.3}}{9.6} & \textbf{66.7} \\
        \bottomrule
    \end{tabular}%
    }
\end{table}

\subsection{Real-Robot Task Definitions}
\label{sec:appendix-real-robot-tasks}

\paragraph{TX-G2.}
\texttt{Cutlery Transfer} requires moving a spoon and a fork individually
between bowls without lifting the bowl itself. \texttt{Bowl Stacking} stacks
the instructed colored bowl. \texttt{Clothes Sorting} places the instructed
garment into a basket. \texttt{Dish Racking} places the instructed plate into
a tight rack.

\begin{figure*}[t]
    \centering
    \setlength{\tabcolsep}{2pt}
    \begin{tabular}{cccc}
        \includegraphics[width=0.235\textwidth]{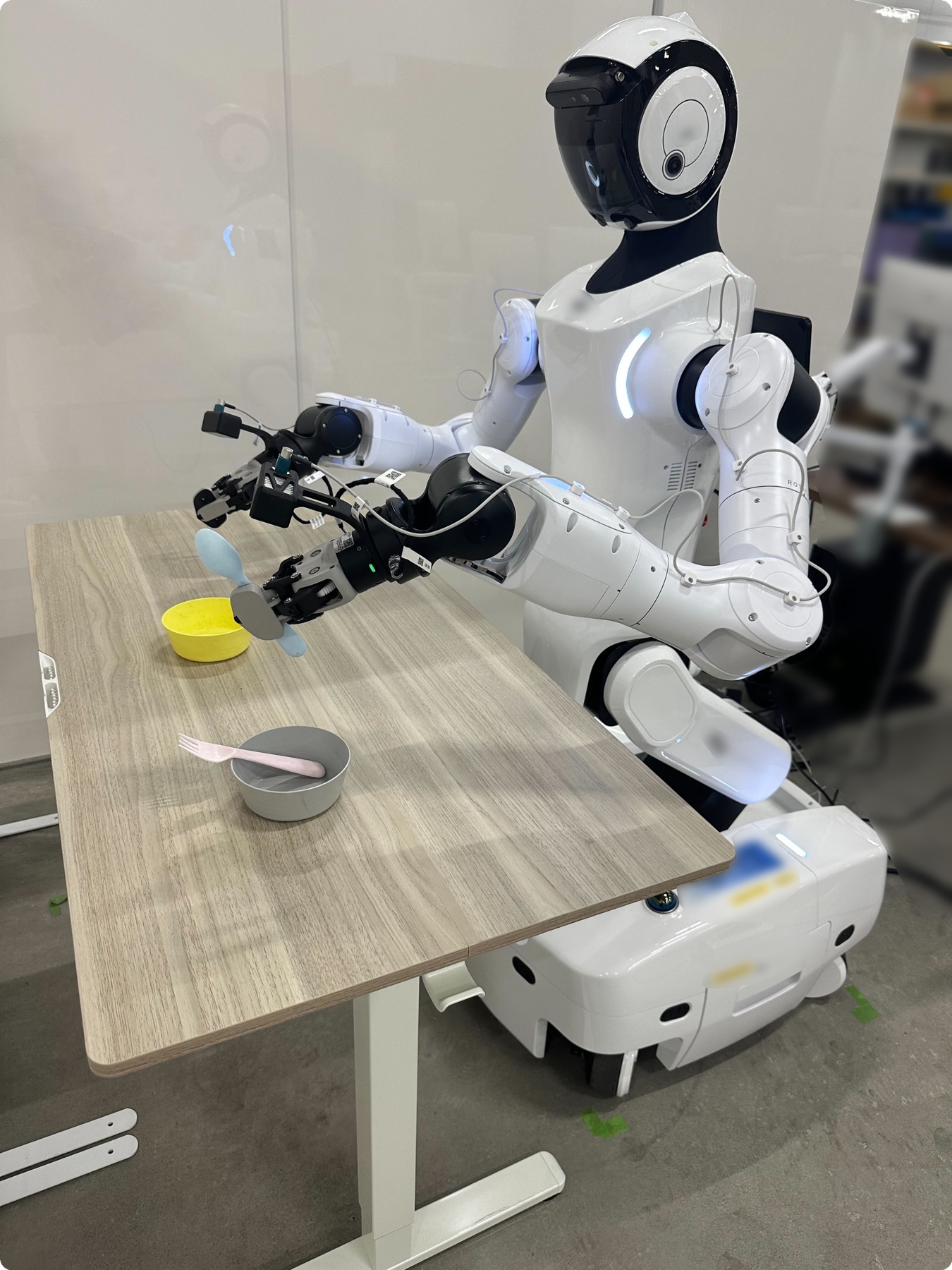} &
        \includegraphics[width=0.235\textwidth]{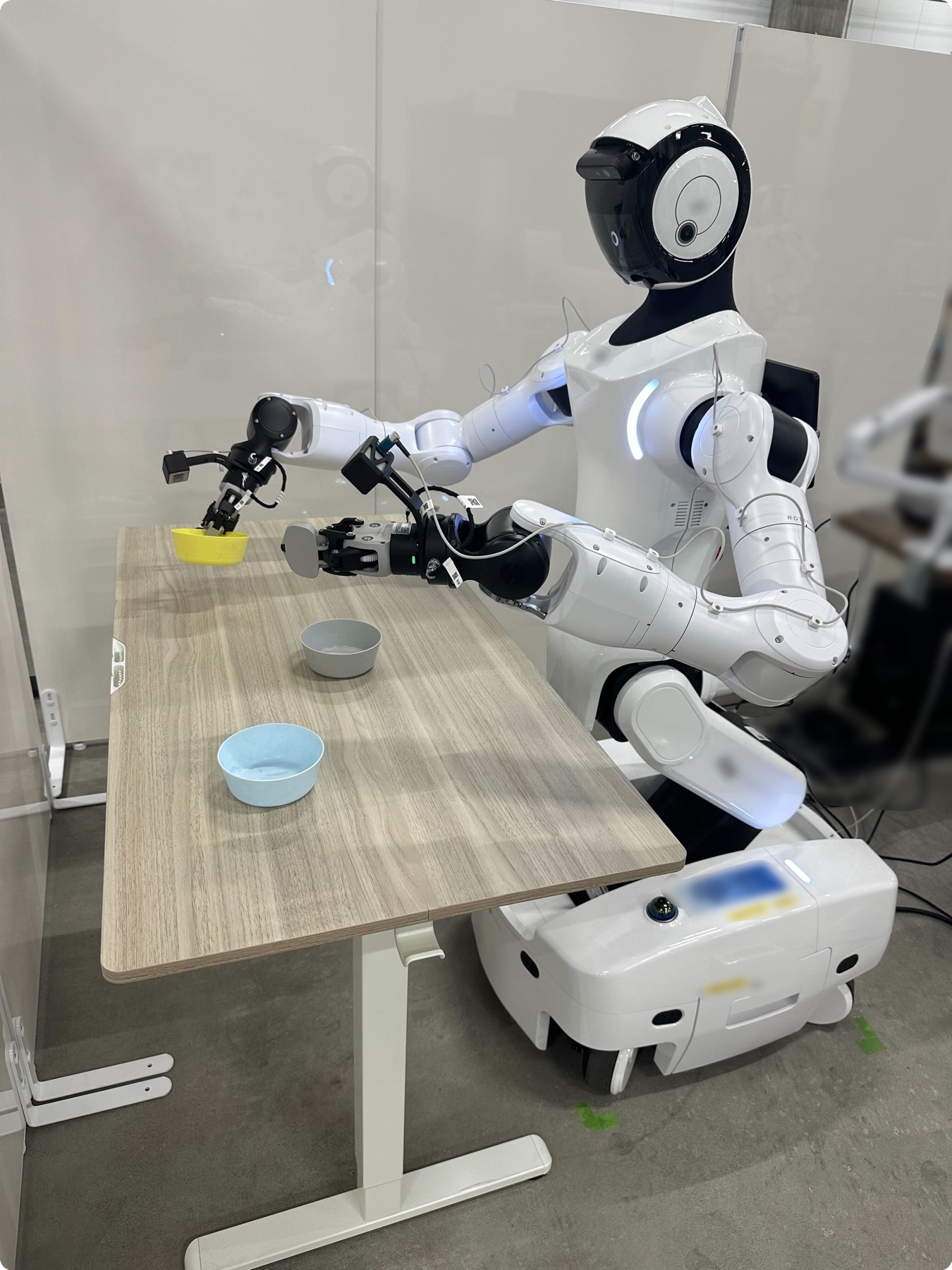} &
        \includegraphics[width=0.235\textwidth]{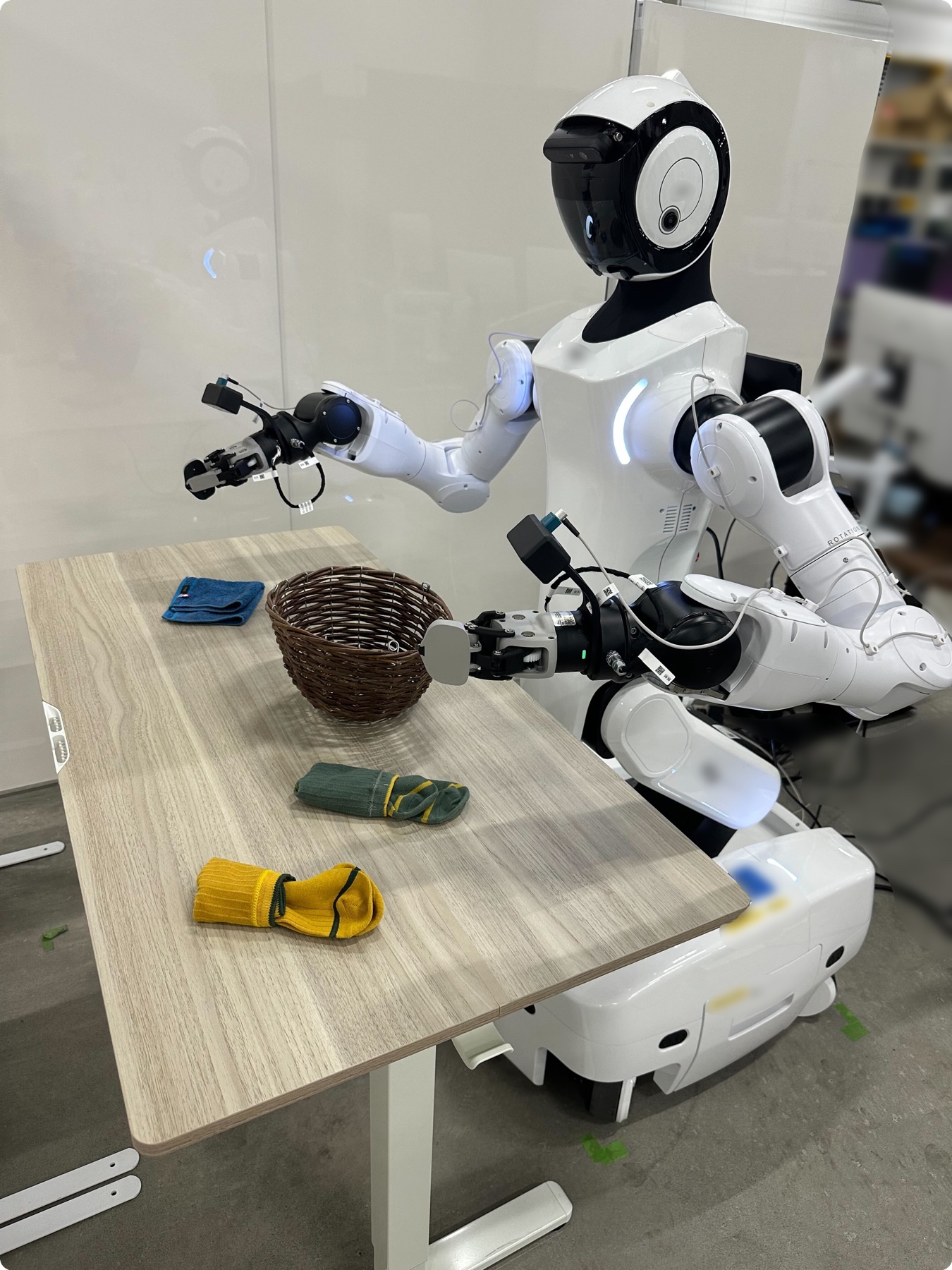} &
        \includegraphics[width=0.235\textwidth]{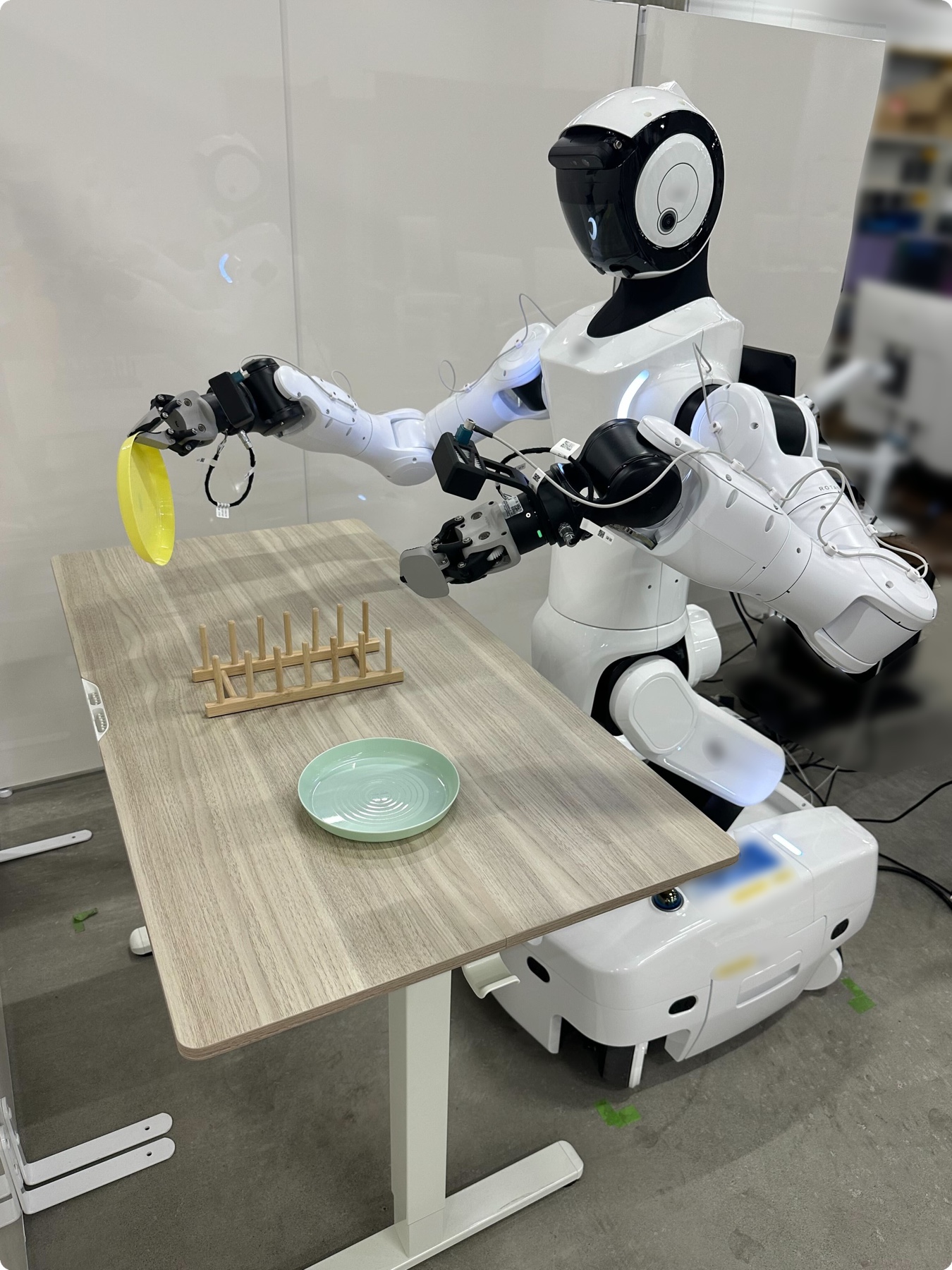} \\
        \scriptsize \textbf{Cutlery Transfer} &
        \scriptsize \textbf{Bowl Stacking} &
        \scriptsize \textbf{Clothes Sorting} &
        \scriptsize \textbf{Dish Racking} \\
    \end{tabular}
    \caption{Representative TX-G2 evaluation trajectories for the four bimanual manipulation tasks used in the appendix real-robot study. Each panel shows keyframes from a successful rollout, highlighting arm selection, semantic object grounding, and precise placement under the TX-G2 benchmark.}
    \label{fig:g2-trajectories}
\end{figure*}

\paragraph{HSR.}
The HSR suite combines locomotion with manipulation. In \texttt{Coffee}, the
robot must navigate to a coffee bottle, grasp it, move to the box labeled
\texttt{1}, and place it through a narrow opening. \texttt{Bottles} consists
of bottle-transport tasks under similar mobile-manipulation constraints.
\texttt{Box} requires transporting the box labeled \texttt{2} beside the box
labeled \texttt{1}, which is challenging because the box is large and
relatively heavy for the gripper. \texttt{Mug} relocates a mug to the corner
that does not already contain one.

\begin{figure*}[t]
    \centering
    \setlength{\tabcolsep}{4pt}
    \begin{tabular}{cc}
        \includegraphics[width=0.47\textwidth]{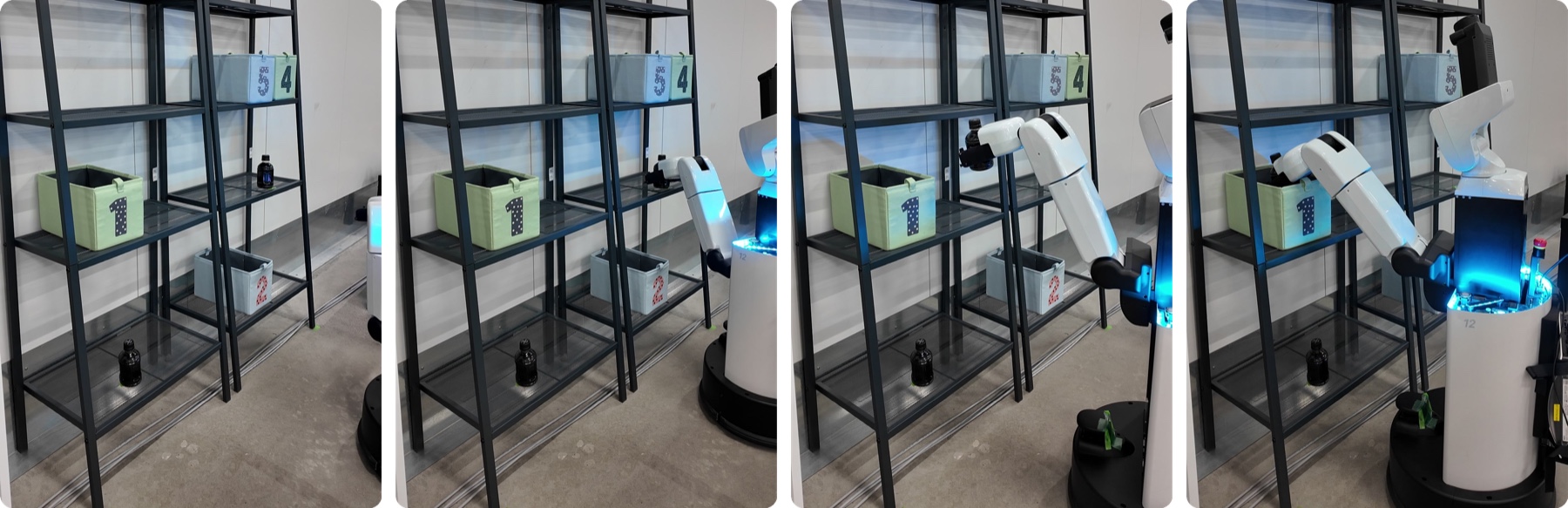} &
        \includegraphics[width=0.47\textwidth]{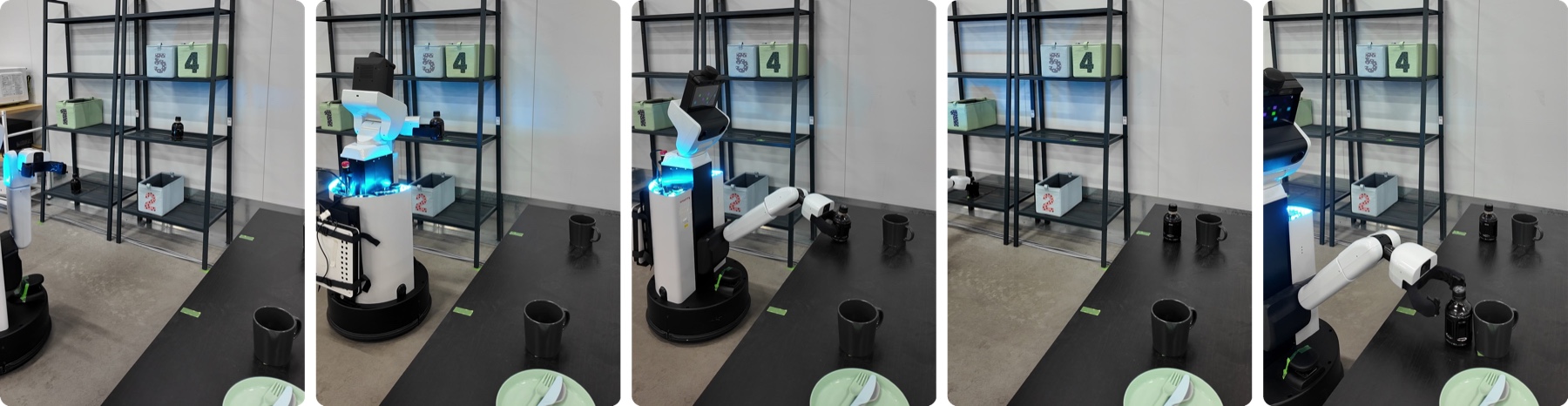} \\
        \scriptsize \textbf{Coffee Bottle $\rightarrow$ Box} &
        \scriptsize \textbf{Coffee Bottles $\rightarrow$ Table} \\
        \includegraphics[width=0.47\textwidth]{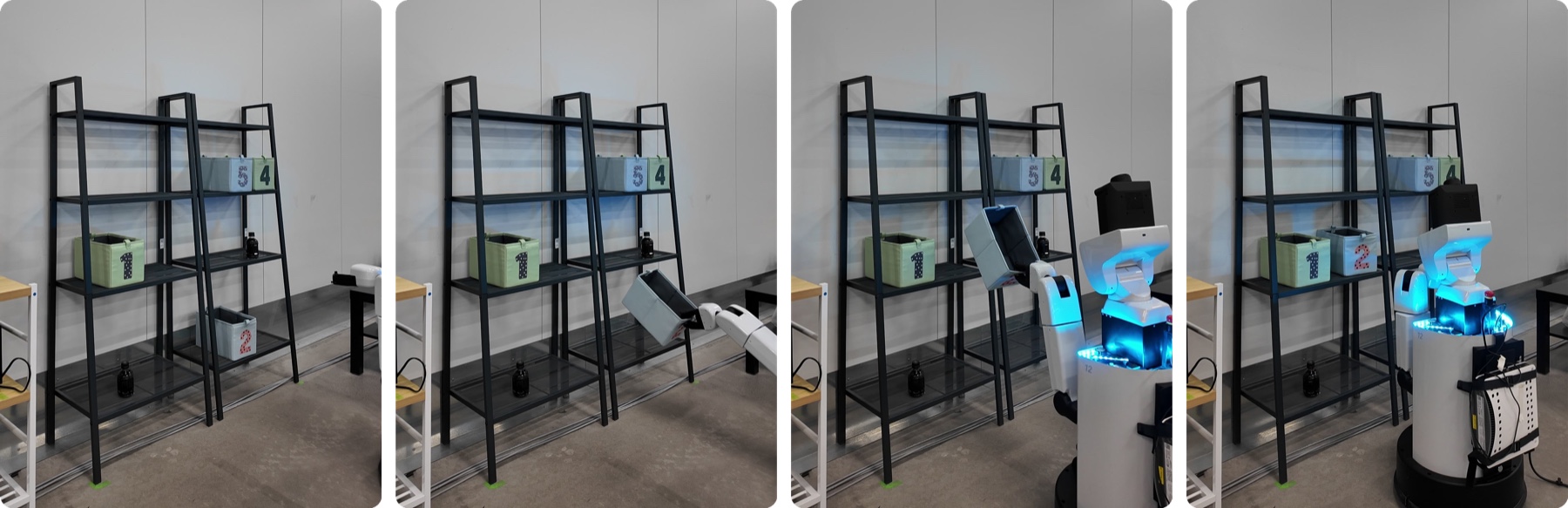} &
        \includegraphics[width=0.47\textwidth]{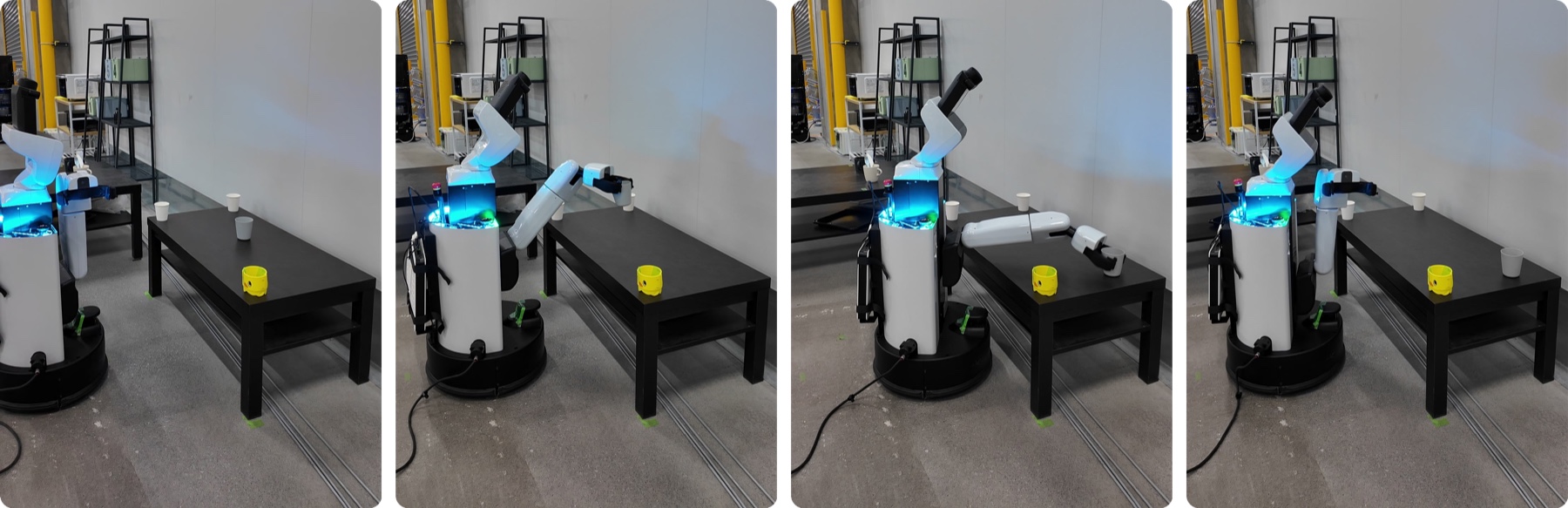} \\
        \scriptsize \textbf{Box Rearrangement} &
        \scriptsize \textbf{Mug Rectangle} \\
    \end{tabular}
    \caption{Representative HSR evaluation trajectories for the four mobile-manipulation tasks used in the appendix real-robot study. Each panel shows keyframes from a successful rollout, illustrating the combination of locomotion, object grounding, and precise placement required by the HSR benchmark.}
    \label{fig:hsr-trajectories}
\end{figure*}

\subsection{Additional Real-Robot Mask/Attention Comparisons}
\label{sec:appendix-mask-attention}

\begin{figure*}[t]
    \centering
    \setlength{\tabcolsep}{4pt}
    \begin{tabular}{cc}
        \multicolumn{2}{c}{\scriptsize \textbf{TX-G2}} \\
        \includegraphics[width=0.48\textwidth]{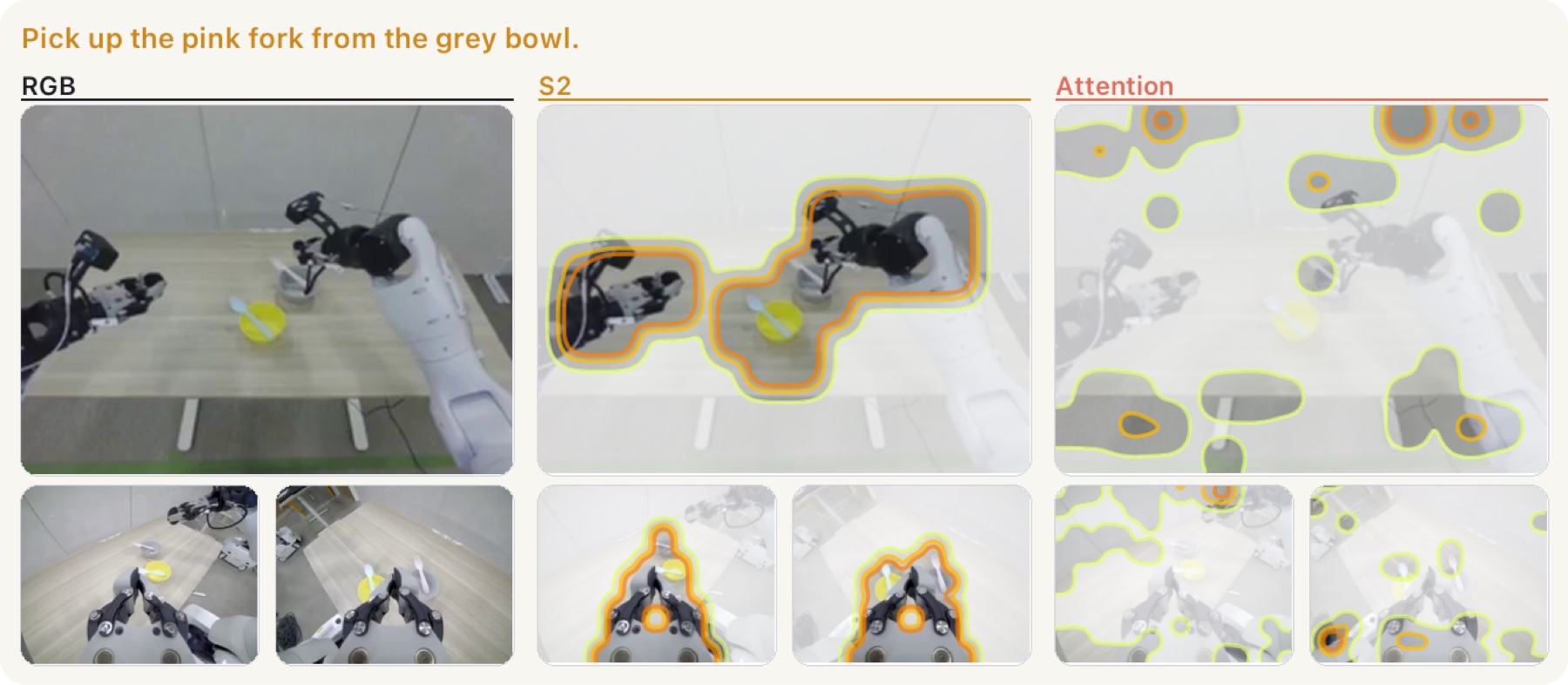} &
        \includegraphics[width=0.48\textwidth]{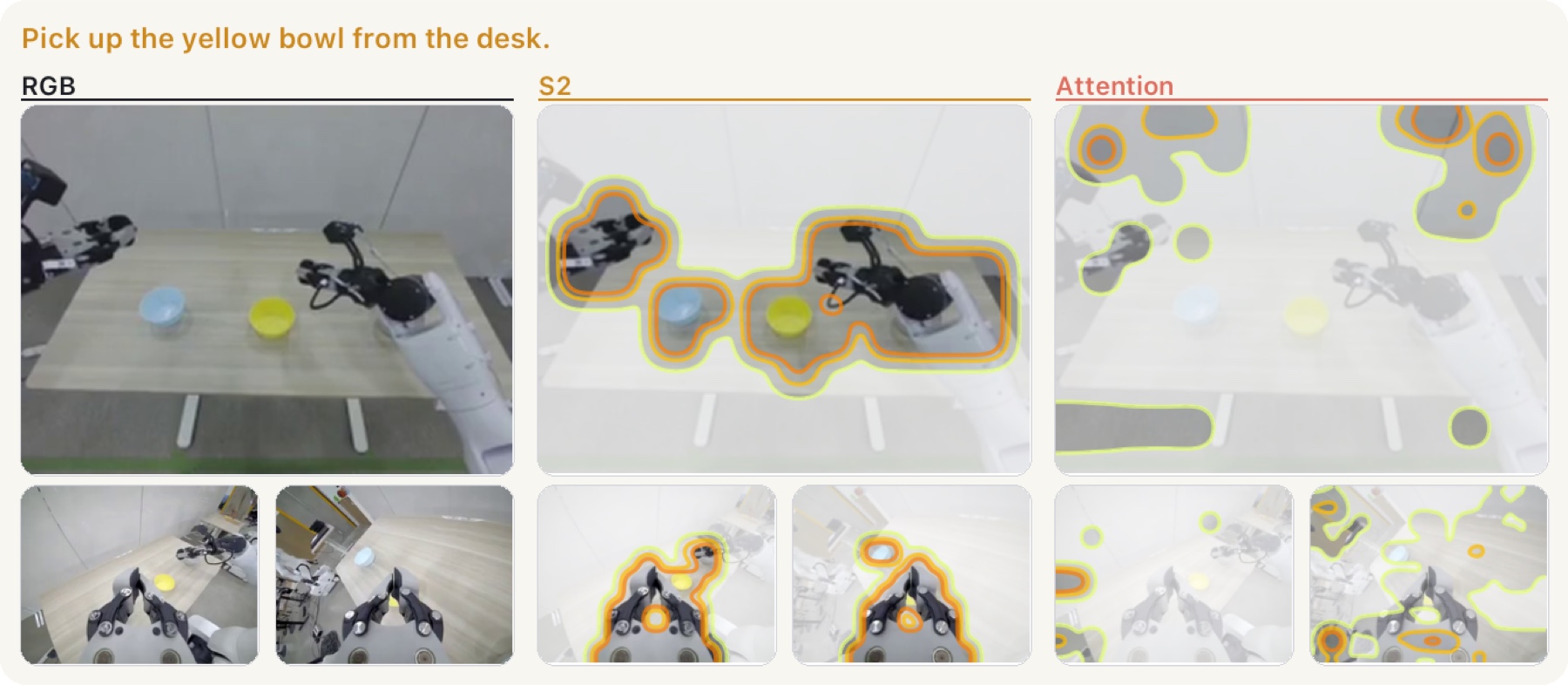} \\
        \includegraphics[width=0.48\textwidth]{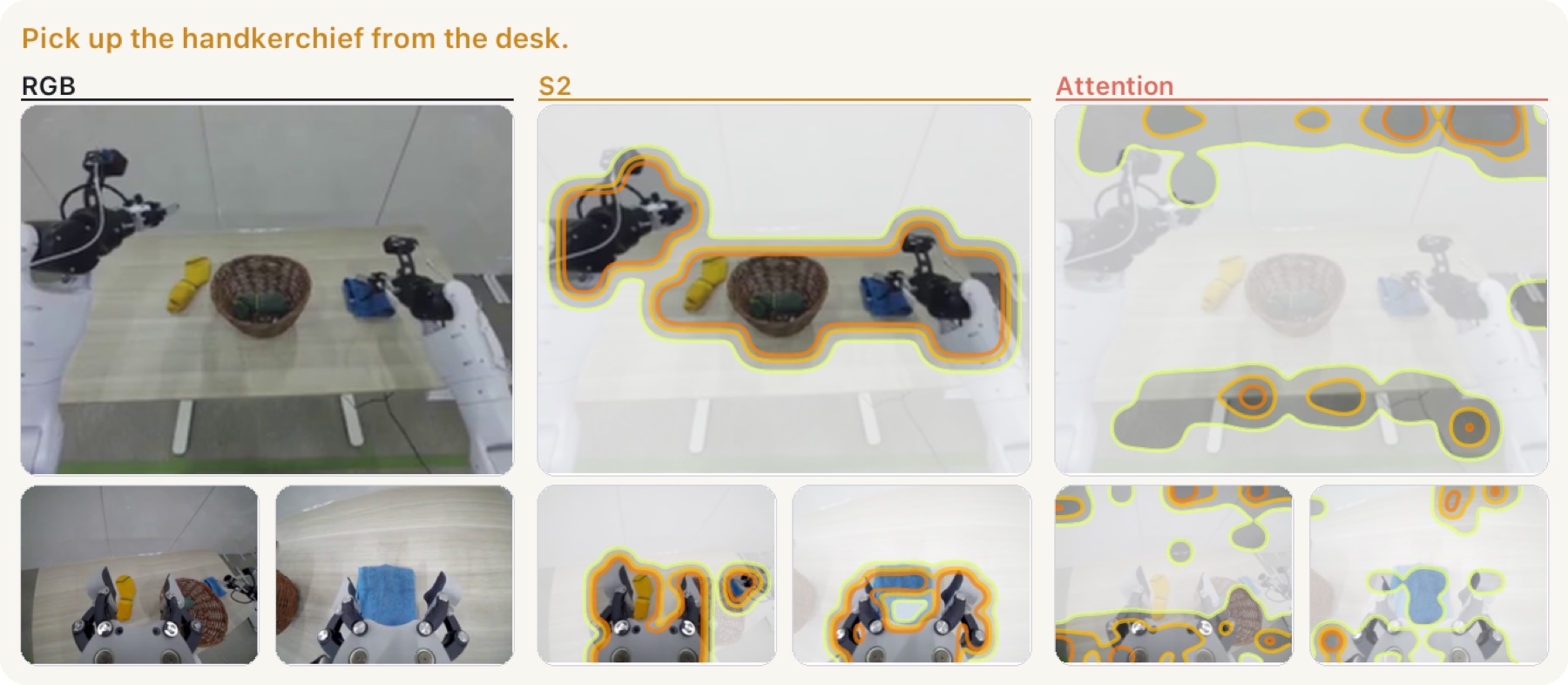} &
        \includegraphics[width=0.48\textwidth]{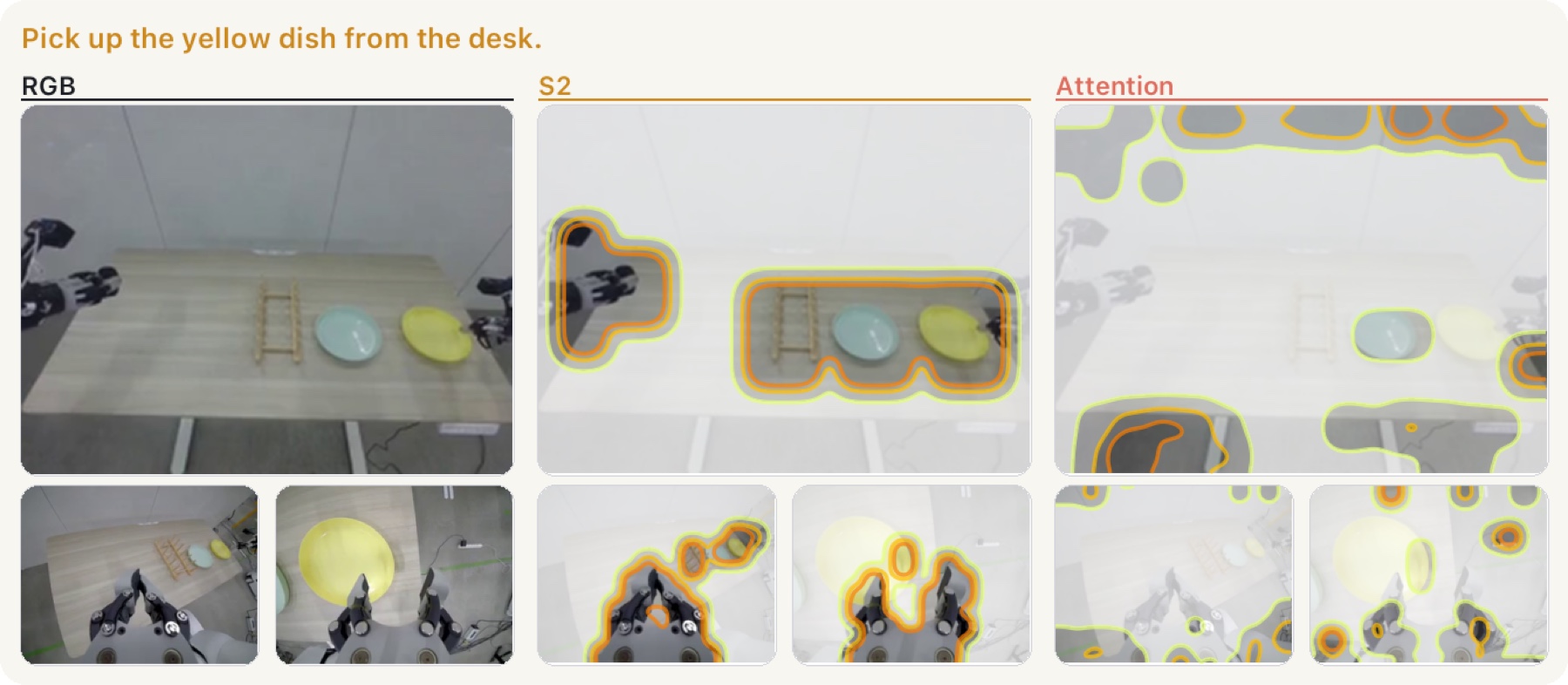} \\
        \multicolumn{2}{c}{\scriptsize \textbf{HSR}} \\
        \includegraphics[width=0.48\textwidth]{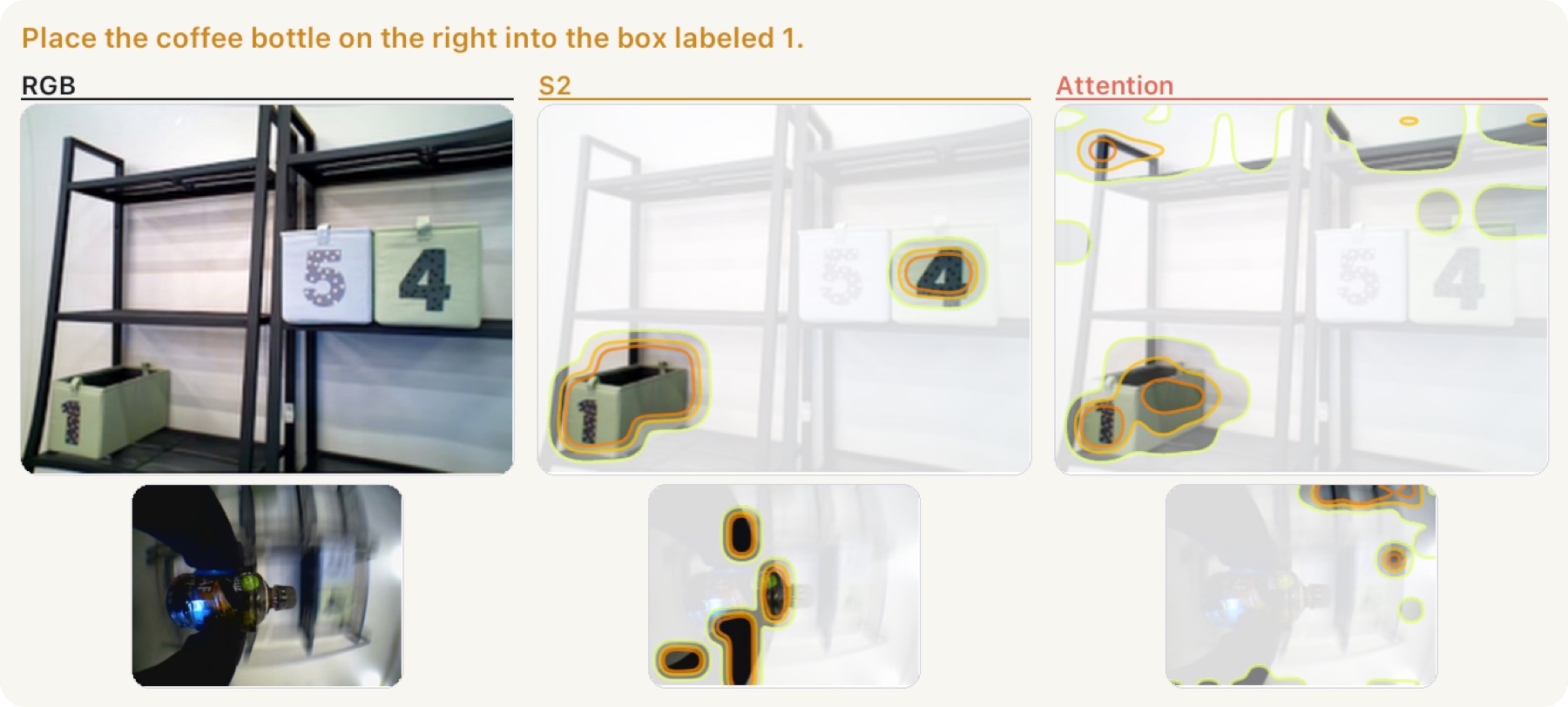} &
        \includegraphics[width=0.48\textwidth]{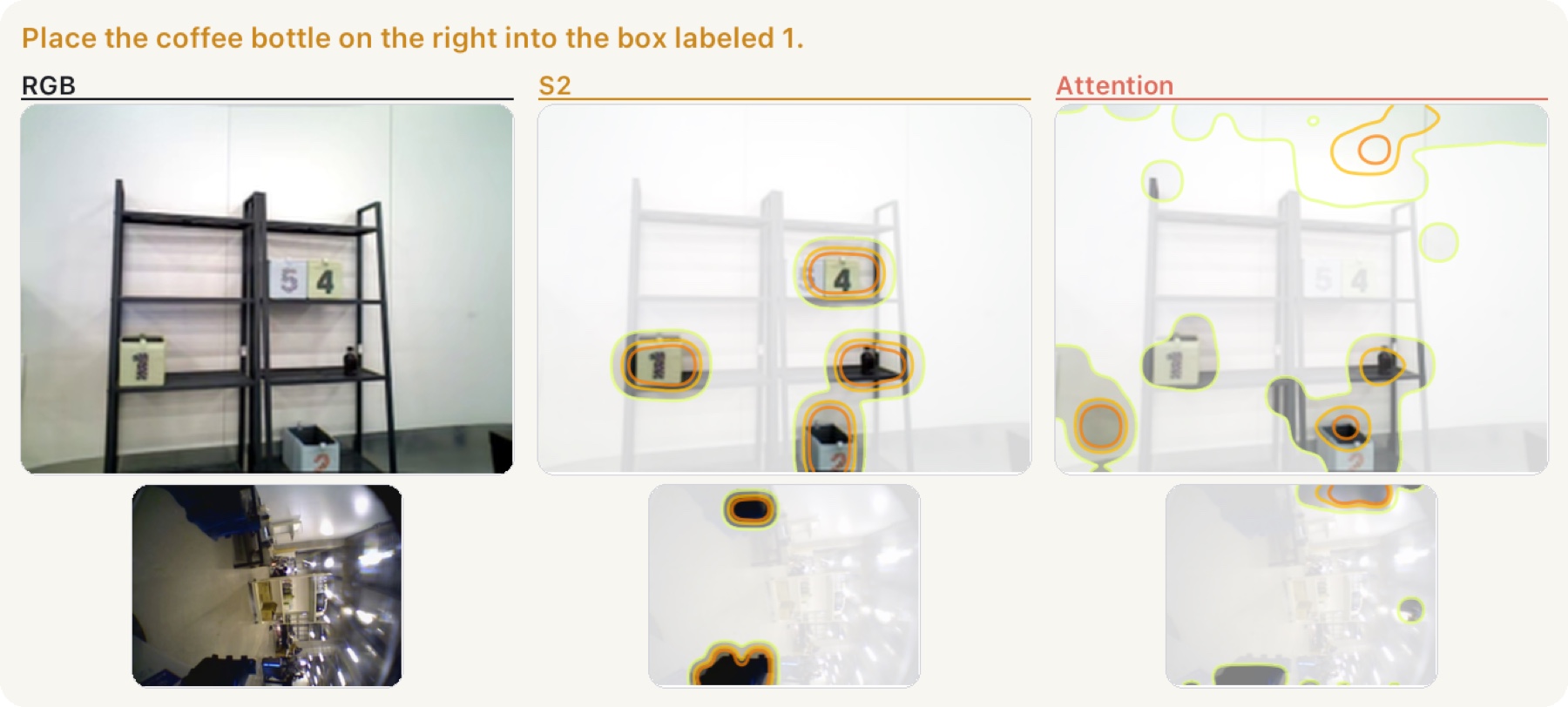} \\
        \includegraphics[width=0.48\textwidth]{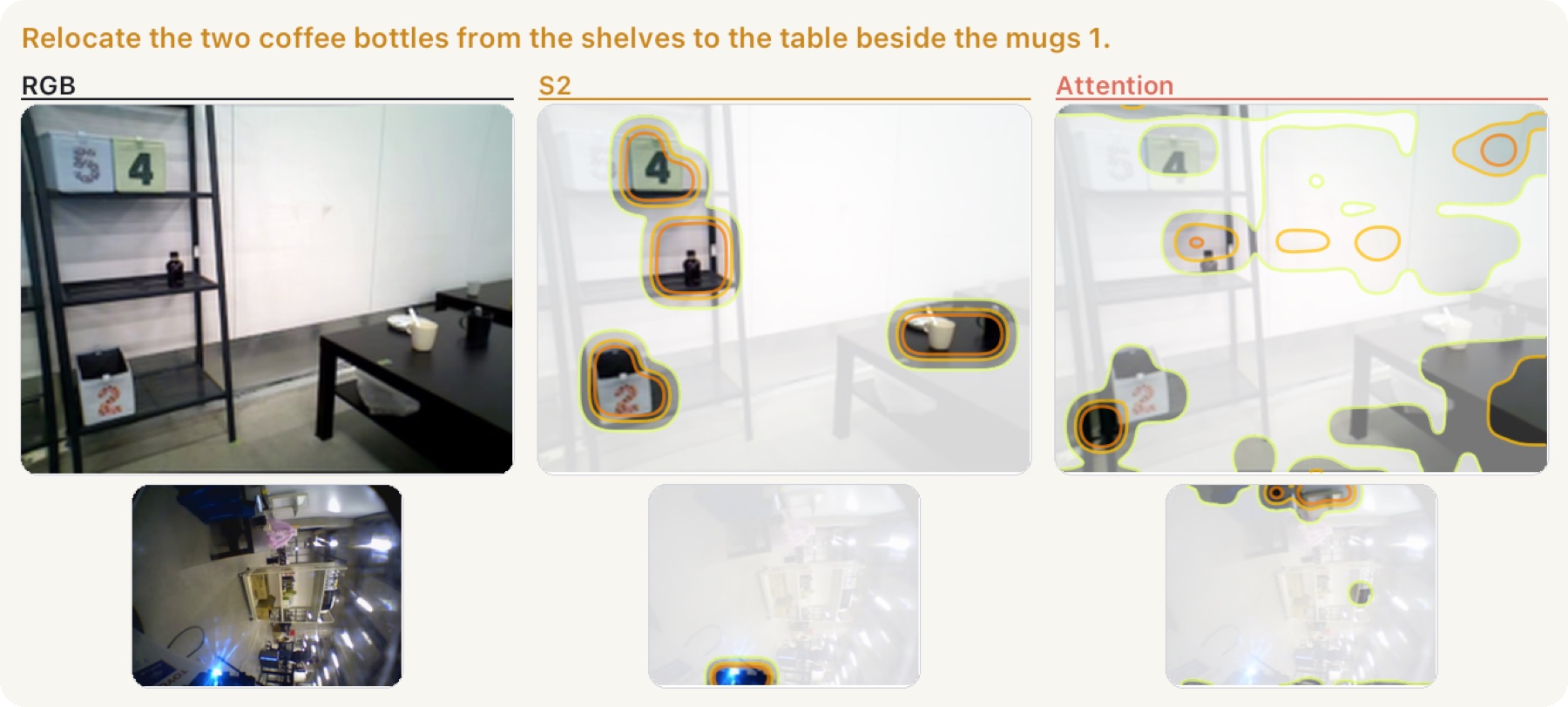} &
        \includegraphics[width=0.48\textwidth]{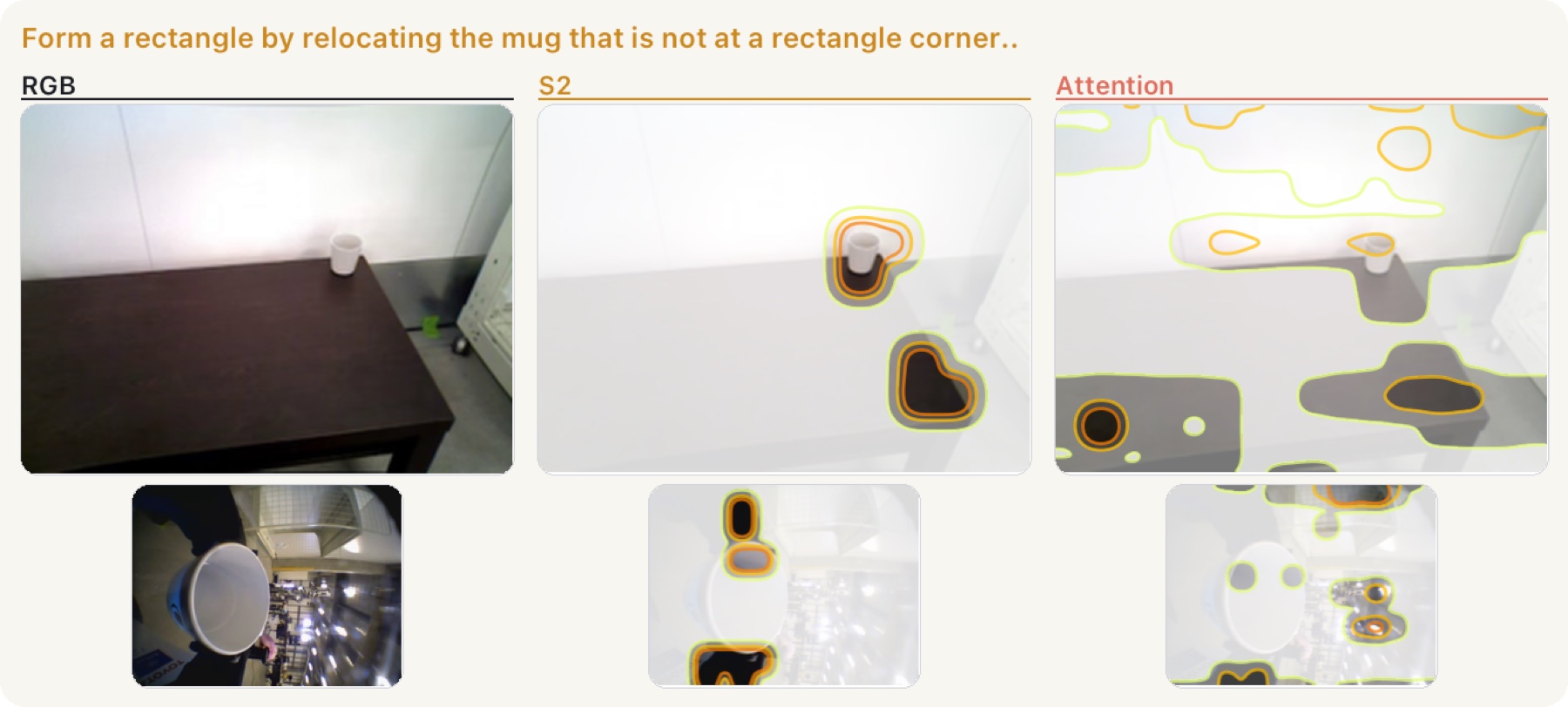} \\
    \end{tabular}
    \caption{Additional real-robot qualitative comparisons between the learned
    visual evidence budget and the backbone's native attention. Each panel
    shows the RGB observation, the learned S2 mask, and the corresponding
    $\pi_{0.5}$ attention map for the same state. Across both TX-G2 and HSR,
    VEB remains more concentrated on the object, contact region, and target
    placement context relevant to the current behavior, while native attention
    is often more diffuse.}
    \label{fig:appendix-mask-attention}
\end{figure*}

Figure~\ref{fig:appendix-mask-attention} expands the qualitative comparison in
the main paper with additional TX-G2 and HSR examples. Across both robots, the
learned VEB masks remain more selective than the backbone's native attention:
they concentrate on the manipulated object, end-effector, and destination
region needed for the current local behavior, while the backbone attention
often spreads across broader scene context or weakly task-related regions. The
examples also illustrate that the retained evidence shifts with task phase,
from source-object grounding to target-region placement, without any region or
mask annotation.

\subsection{Cluttered-Scene Stress Tests}
\label{sec:appendix-cluttered}

We additionally evaluated S2 on a deliberately cluttered TX-G2 clothes-sorting
setup designed to probe robustness beyond the training distribution. Only the
two sock colors, the handkerchief, and the basket were seen during training;
the remaining tabletop objects were unseen distractors. The policy was required
to complete the ordered sequence \texttt{green socks $\rightarrow$ handkerchief
$\rightarrow$ yellow socks}, while a human operator moved distractor objects
and, in some cases, the basket itself during execution. Despite these
interventions and the changing clutter layouts, S2 completed all three tested
rollouts successfully.

\begin{figure*}[t]
    \centering
    \setlength{\tabcolsep}{8pt}
    \begin{tabular}{cc}
        \includegraphics[width=0.34\textwidth]{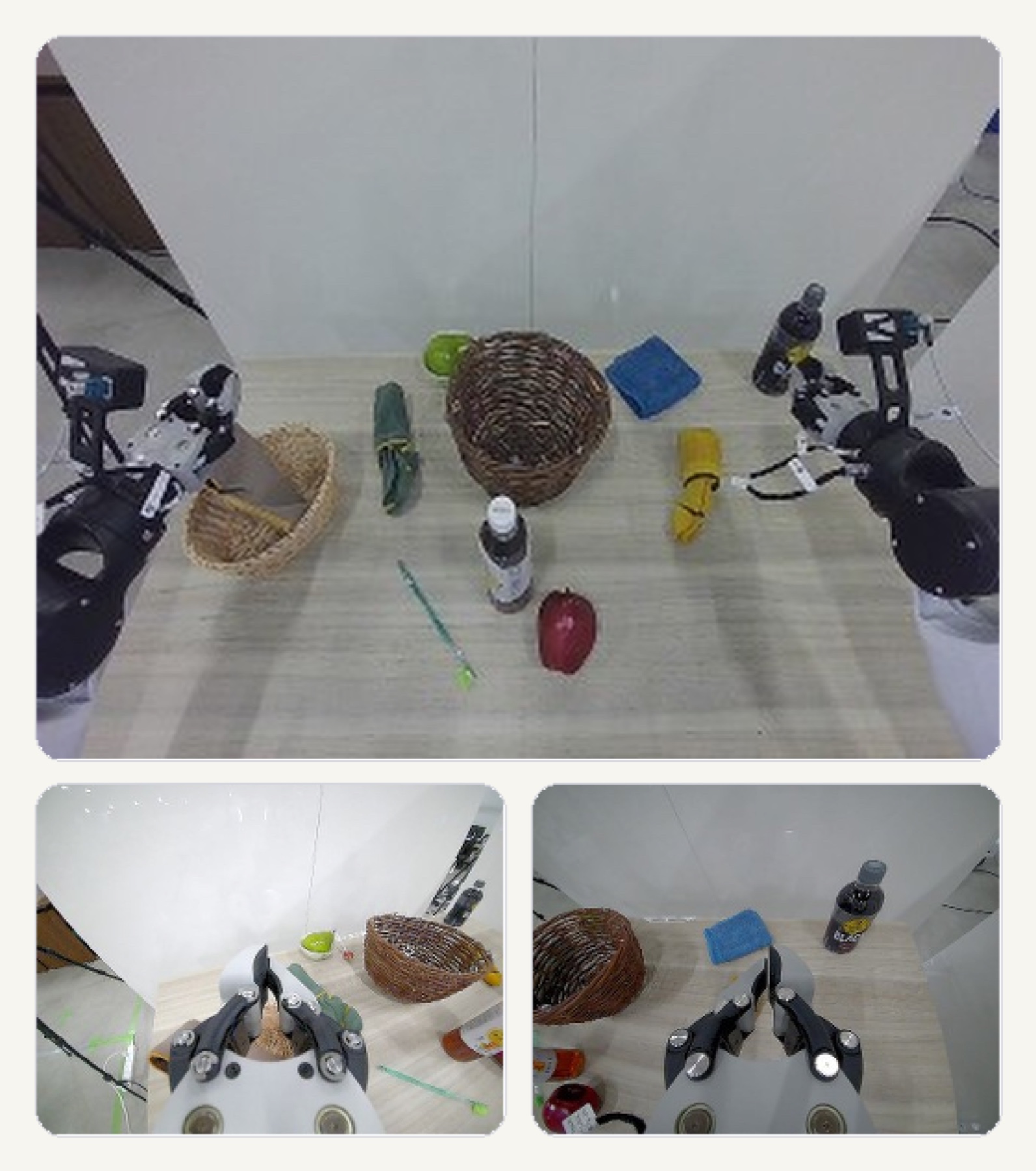} &
        \includegraphics[width=0.34\textwidth]{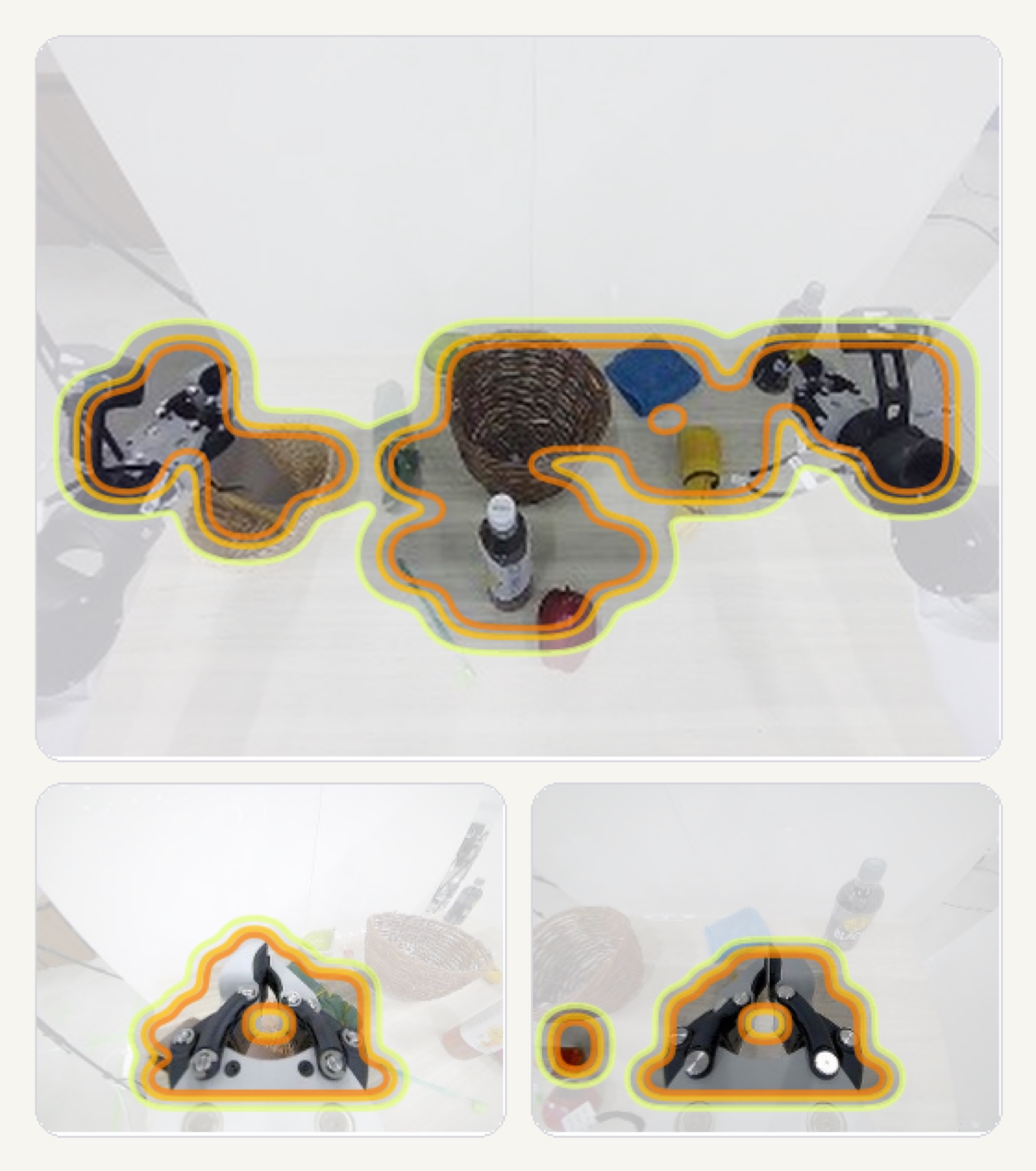} \\
    \end{tabular}
    \caption{Representative cluttered TX-G2 observation (left) and the
    corresponding S2 mask (right) for the ordered clothes-sorting task. The
    tabletop contains many distractors absent from the training set, yet the
    learned mask remains concentrated on the task-relevant garment,
    end-effector, and basket region.}
    \label{fig:cluttered-request}
\end{figure*}

\begin{figure*}[t]
    \centering
    \begin{tabular}{c}
        \includegraphics[width=0.94\textwidth]{figures/cluttered/DJI_20260522183028_0388_D_1x4.jpg} \\
        \includegraphics[width=0.94\textwidth]{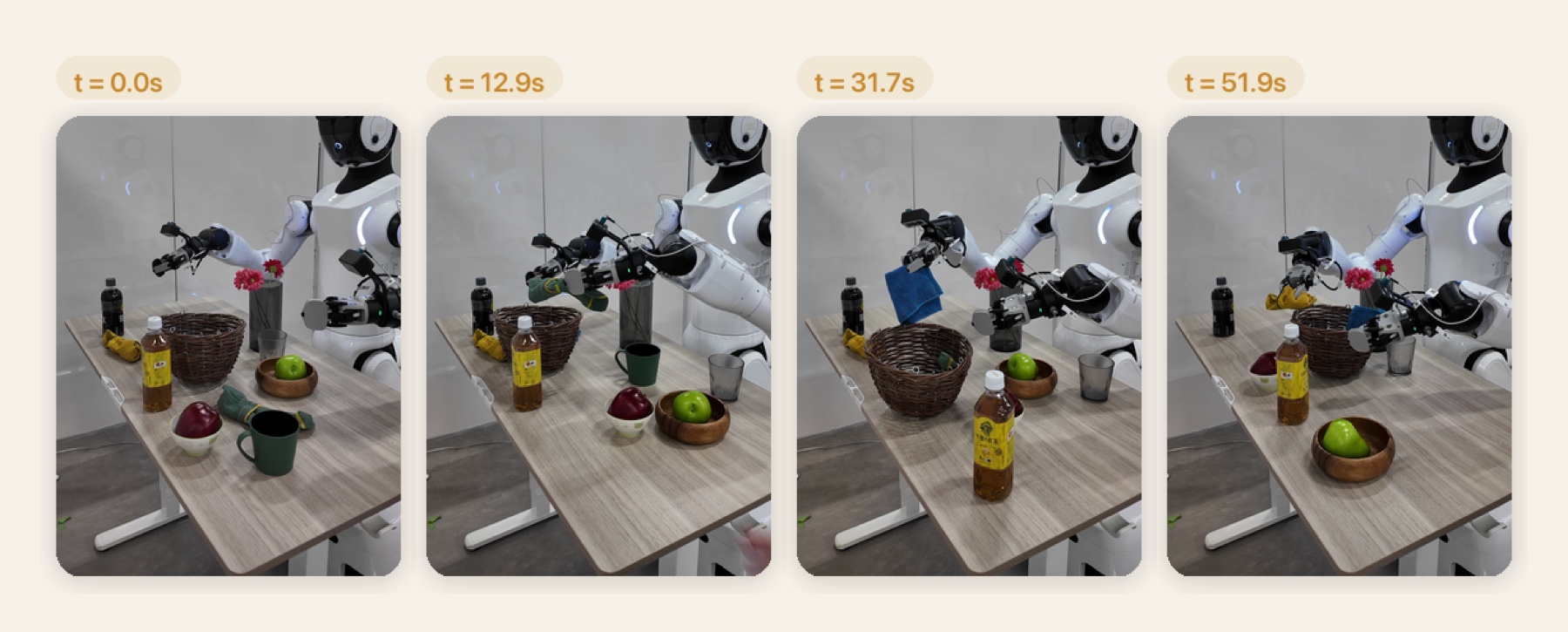} \\
        \includegraphics[width=0.94\textwidth]{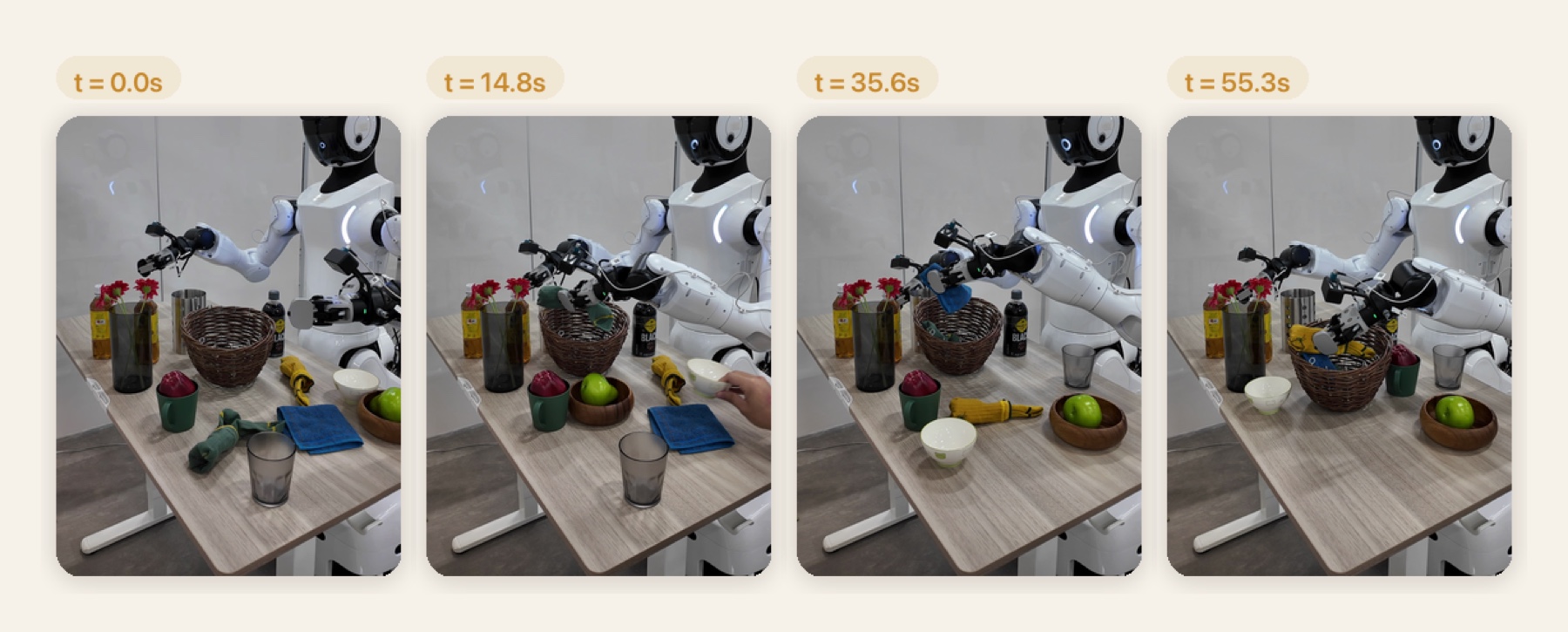} \\
    \end{tabular}
    \caption{Three successful TX-G2 cluttered-scene rollouts for the ordered
    sequence \texttt{green socks $\rightarrow$ handkerchief $\rightarrow$
    yellow socks}. Each row shows a different initial clutter layout with
    different unseen distractors. During execution, a person also perturbs
    object or basket positions, forcing the policy to re-ground the current
    target online while preserving the required completion order.}
    \label{fig:cluttered-rollouts}
\end{figure*}

Figure~\ref{fig:cluttered-request} shows a representative cluttered observation
and the corresponding S2 mask. Even in the presence of many unseen objects, the
learned mask concentrates on the task-relevant garment, end-effector, and
placement region rather than diffusing across the full scene. Figure~\ref{fig:cluttered-rollouts}
then shows three successful rollouts under different initial clutter
arrangements and intervention patterns, illustrating that the policy can
re-ground the current target and continue the required ordering constraint.

\subsection{Why VLA-Adapter Underperforms on TX-G2}
\label{sec:appendix-real-robot-analysis}

We did not find evidence of a major inference-time integration failure for
VLA-Adapter. Its preprocessing path, proprio normalization, and bundle-level
I/O all passed our validation checks. However, in a 30-sample dataset-driven
replay test, its predicted actions were only marginally better than a
mean-action baseline, suggesting that the fine-tuned policy remained close to
an average regressor rather than a strongly task-conditioned controller. This
diagnosis is consistent with its 0\% closed-loop success on TX-G2.

\subsection{Full Core Ablation Breakdown}
\label{sec:appendix-core-ablation}

Table~\ref{tab:core-object-goal-full} reports the full perturbation breakdown
for the mean-only comparison shown in the main text.

\begin{table}[h]
    \centering
    \scriptsize
    \setlength{\tabcolsep}{2.7pt}
    \caption{Full perturbation breakdown for Figure~\ref{fig:core-object-goal-means}.
    \emph{Original} denotes the original-instruction baseline; all other rows
    are S2 ablations.}
    \label{tab:core-object-goal-full}
    \renewcommand{\arraystretch}{1.18}
    \adjustbox{max width=\textwidth}{%
    \begin{tabular}{>{\raggedright\arraybackslash}m{2.2cm}|*{5}{>{\centering\arraybackslash}m{0.72cm}}|*{5}{>{\centering\arraybackslash}m{0.72cm}}}
        \toprule
        \multirow{2}{2.2cm}{\textbf{Methods}} & \multicolumn{5}{c|}{\texttt{libero\_goal}} & \multicolumn{5}{c}{\texttt{libero\_object}} \\
        & Obj. & Pos. & Sem. & Task & Mean & Obj. & Pos. & Sem. & Task & Mean \\
        \midrule
        Original & 90.0 & 29.0 & 95.0 & 17.0 & 57.8 & 89.0 & 19.0 & 95.0 & 10.0 & 53.3 \\
        Refined & 56.0 & 24.0 & 61.0 & 10.0 & 37.8 & 78.0 & \textbf{69.5} & 98.5 & 35.5 & 70.4 \\
        Original + VEB & 87.0 & 36.0 & \textbf{97.0} & 30.0 & 62.5 & \textbf{93.0} & 33.5 & 99.5 & 21.0 & 61.8 \\
        Refined + VEB & 52.0 & 18.0 & 60.0 & 10.0 & 35.0 & 88.0 & 68.0 & 99.5 & 32.0 & 71.9 \\
        Hybrid, no VEB & 56.5 & 26.0 & 58.5 & 10.5 & 37.9 & 73.0 & 48.5 & 98.5 & 28.5 & 62.1 \\
        \rowcolor{TableHighlight}
        \textbf{S2} & \textbf{90.5} & \textbf{42.0} & 95.5 & \textbf{52.5} & \textbf{70.1} & 90.5 & 64.5 & \textbf{100.0} & \textbf{37.5} & \textbf{73.1} \\
        \bottomrule
    \end{tabular}
    }
\end{table}